\newcommand{\status}{1}

\newcommand{\Description}[1]{#1}

\documentclass[10pt,twocolumn,letterpaper]{article}

\usepackage{cvpr}              

\definecolor{cvprblue}{rgb}{0.21,0.49,0.74}
\usepackage[pagebackref,breaklinks,colorlinks,allcolors=cvprblue]{hyperref}

\input{conf/packages}
\definecolor{INCOMPLETECOLOR}{RGB}{178,34,34}
\definecolor{UNDERREVISIONCOLOR}{RGB}{210,121,121}
\definecolor{FEEDBACKNEEDEDCOLOR}{RGB}{230,170,50}
\definecolor{FEEDBACKGIVENCOLOR}{RGB}{121,210,121}
\definecolor{COMPLETECOLOR}{RGB}{121,124,210}
\definecolor{QINCHANCOLOR}{RGB}{0,190,255}

\definecolor{TODOCOLOR}{RGB}{255,0,0}
\definecolor{MONDECOLOR}{RGB}{0,0,255}
\definecolor{QICOLOR}{RGB}{118,185,0}
\definecolor{ANJULCOLOR}{RGB}{127,127,0}
\definecolor{RACHELCOLOR}{RGB}{127,0,127}
\definecolor{KENNYCOLOR}{RGB}{255,127,127}
\definecolor{GUESTCOLOR}{RGB}{0,127,127}

\definecolor{WHITE}{RGB}{255,255,255}

\newcommand{\nothing}[1]{}
\newcommand{\isolated}[1]{\hfill\break#1\xspace}
\newcommand{\Caption}[2]{\caption[#1]{{\em #1} #2}}

\ifthenelse{\equal{\status}{0}} {                     

    \newcommand{\todo}[1]{%
        \addcontentsline{toc}{subsection}{
            \protect\numberline{}
            \textcolor{TODOCOLOR}{[TODO] #1}}
            \textcolor{TODOCOLOR}{[TODO] \emph{#1}}}
    \newcommand{\warning}[1]{\todo{#1}}
    \newcommand{\note}[1]{{\it\color{blue} #1}}

    \newcommandx{\monde}[2][1=]
        {\setulcolor{MONDECOLOR}{\ul{#1}}
         \isolated{\textcolor{MONDECOLOR}{\textbf{Monde:} #2}}}
    \newcommandx{\qisun}[2][1=]
        {\setulcolor{QICOLOR}{\ul{#1}}
         \isolated{\textcolor{QICOLOR}{\textbf{Qi:} #2}}}
    \newcommandx{\anjul}[2][1=]
        {\setulcolor{ANJULCOLOR}{\ul{#1}}
         \isolated{\textcolor{ANJULCOLOR}{\textbf{Anjul:} #2}}}
    \newcommandx{\rachel}[2][1=]
        {\setulcolor{RACHELCOLOR}{\ul{#1}}
         \isolated{\textcolor{RACHELCOLOR}{\textbf{Rachel:} #2}}}
    \newcommandx{\qinchan}[2][1=]
        {\setulcolor{QINCHANCOLOR}{\ul{#1}}
         \isolated{\textcolor{QINCHANCOLOR}{\textbf{Qinchan (Wing):} #2}}}
    \newcommandx{\kenny}[2][1=]
        {\setulcolor{KENNYCOLOR}{\ul{#1}}
         \isolated{\textcolor{KENNYCOLOR}{\textbf{Kenny:} #2}}}

    \newcommandx{\guest}[3][1=]
        {\setulcolor{GUESTCOLOR}{\ul{#1}} \textcolor{GUESTCOLOR}
        {[\textbf{#2:} #3]}}
}{                                                    
    \newcommand{\todo}[1]{}
    \newcommand{\warning}[1]{}
    \newcommand{\note}[1]{}

    \newcommandx{\monde}[2][1=]{#1}
    \newcommandx{\qinchan}[2][1=]{#1}
    \newcommandx{\qisun}[2][1=]{#1}
    \newcommandx{\kenny}[2][1=]{#1}
    \newcommandx{\guest}[3][1=]{#1}
}

\ifthenelse{\equal{\status}{0}} {          
    
}{}
\ifthenelse{\equal{\status}{1}} {          
    
}{}
\ifthenelse{\equal{\status}{2}} {          
    
}{}
\ifthenelse{\equal{\status}{3}} {          
    
}{}
\ifthenelse{\equal{\status}{4}} {          
    
}{}

\ifthenelse{\equal{\status}{0}} {
    \newcommand{\badge}[2]{\colorbox{#1}{\small\textcolor{WHITE}{\texttt{#2}}}}
    \newcommand{\headerBadge}[2]{\hspace*{\fill}\badge{#1}{#2}}
    
    \newcommand{\complete}{\headerBadge{COMPLETECOLOR}{complete}}
    \newcommand{\feedbackGiven}{\headerBadge{FEEDBACKGIVENCOLOR}{feedback given}}
    \newcommand{\feedbackNeeded}{\headerBadge{FEEDBACKNEEDEDCOLOR}{feedback needed}}
    \newcommand{\underRevision}{\headerBadge{UNDERREVISIONCOLOR}{under revision}}
    \newcommand{\incomplete}{\headerBadge{INCOMPLETECOLOR}{incomplete}}
}{
    \newcommand{\badge}[2]{}{}
    \newcommand{\headerBadge}[2]{}{}
    
    \newcommand{\complete}{}
    \newcommand{\feedbackGiven}{}
    \newcommand{\feedbackNeeded}{}
    \newcommand{\underRevision}{}
    \newcommand{\incomplete}{}
}

\newcommand{\lstm}{\theta}
\newcommand{\mlp}{MLP}
\newcommand{\sigmoid}{Sigmoid}

\newcommand{\scores}{scores}
\newcommand{\GaussianKernal}{G}

\newcommand{\image}{I}

\newcommand{\SNRLaplacian}{L-SNR\xspace}
\newcommand{\dreamsim}{D-SIM\xspace}
\newcommand{\dreamsimscore}{D_s}
\newcommand{\clip}{I-CLIP\xspace}

\newcommand{\prompt}{p}

\newcommand{\selectedPromptSet}{\mathbf{P}}

\newcommand{\CLIPembedding}[1]{\overline{#1}}
\newcommand{\positionEmbedding}[1]{\overline{#1}}
\newcommand{\CLIPcaptioning}{C}
\newcommand{\CLIPsimilarity}{C_{clip}}
\newcommand{\thresholdClip}{\hat{\similarity}}

\newcommand{\step}{t}
\newcommand{\stepOptimal}{\step^{*}}
\newcommand{\stepSetSize}{N}
\newcommand{\stepSet}{\mathbf{T}}
\newcommand{\weightPlateau}{\omega}

\newcommand{\metric}{m}

\newcommand{\flops}{\eta}

\newcommand{\similarity}{S}

\newcommand{\methodName}{BudgetFusion\xspace}

\newcommand{\condOurs}{\textbf{OURS}\xspace}
\newcommand{\condUniform}{\textbf{UNIFORM}\xspace}
\newcommand{\condRef}{\textbf{REFERENCE}\xspace}

\newcommand{\mean}{\mu}
\newcommand{\std}{\sigma}

\usepackage{mathtools}

\usepackage{tikz}
\usetikzlibrary{fit,shapes.geometric}

\def\confName{CVPR}
\def\confYear{2025}

\title{\methodName: Perceptually-Guided Adaptive Diffusion Models}

\author{Qinchan (Wing) Li\footnote{1}  \quad Kenneth Chen\footnotemark[1] \quad Changyue (Tina) Su \quad Qi Sun\\
Tandon School of Engineering, New York University \\
{\tt\small \{ql840, kc4906, cs7483, qisun\}@nyu.edu} \\ 
}

\begin{document}

\twocolumn[{%
\renewcommand\twocolumn[1][]{#1}%
\maketitle
\includegraphics[width=\linewidth]{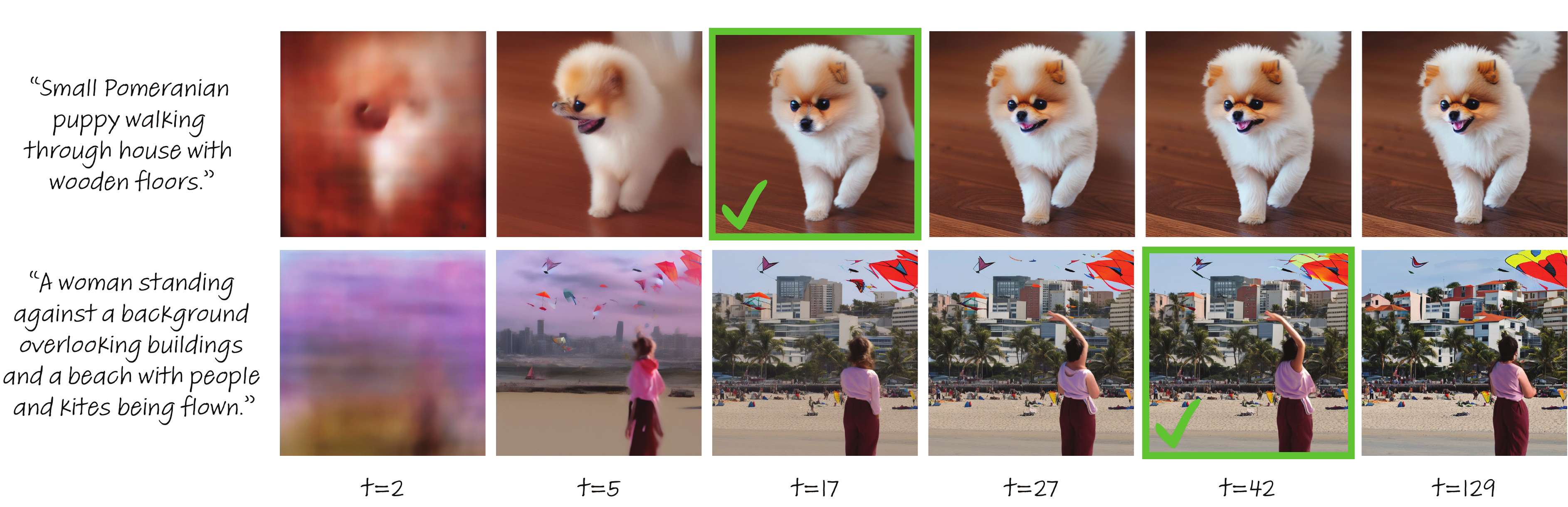}
\captionof{figure}{Given an input text prompt, our \methodName model guides the number of denoising steps $\step$ before the generation starts.  It balances the trade-off between visual perception quality and computational cost, achieving an optimized ``quality gain per denoising step'' efficiency. \vspace{1em}}
\label{fig:teaser}
}]
\begin{abstract}
Diffusion models have shown unprecedented success in the task of text-to-image generation. While these models are capable of generating high-quality and realistic images, the complexity of sequential denoising has raised societal concerns regarding high computational demands and energy consumption. In response, various efforts have been made to improve inference efficiency. However, most of the existing efforts have taken a fixed approach with neural network simplification or text prompt optimization.

Are the quality improvements from all denoising computations equally perceivable to humans? We observed that images from different text prompts may require different computational efforts given the desired content. The observation motivates us to present \methodName, a novel model that suggests the most perceptually efficient number of diffusion steps before a diffusion model starts to generate an image. This is achieved by predicting multi-level perceptual metrics relative to diffusion steps. With the popular Stable Diffusion as an example, we conduct both numerical analyses and user studies. Our experiments show that \methodName saves up to five seconds per prompt without compromising perceptual similarity. We hope this work can initiate efforts toward answering a core question: how much do humans perceptually gain from images created by a generative model, per watt of energy?
\end{abstract}



\renewcommand*{\thefootnote}{\fnsymbol{footnote}}
\footnotetext[1]{The authors contribute equally to this paper.}
\renewcommand*{\thefootnote}{\arabic{footnote}}

\section{Introduction \incomplete}
\label{sec:introduction}

\note{
What problem we are trying to solve.
Why it is important, and why people should care.
}


Diffusion models have been shown to produce images and videos of increasingly high fidelity \cite{ho2020denoising}. 
However, this capability comes at the cost of significant computational demands due to the large number of inference steps required to generate images of high quality, which each require evaluation of large neural networks commonly having over 1 billion parameters \cite{rombach2022high}. 
The substantial computational burden of diffusion models has prohibited on-device deployment and led to societal concerns about their energy consumption and impact on the environment \cite{genaiEnergy,genaiEnergyNature,kaack2022aligning}.

\note{
What prior works have done, and why they are not adequate.
(Note: this is just high level big ideas. Details should go to a previous work section.)
}

\begin{figure}
\centering
\subfloat[\textit{``a white and empty wall''}]{
    \includegraphics[width=0.47\linewidth]{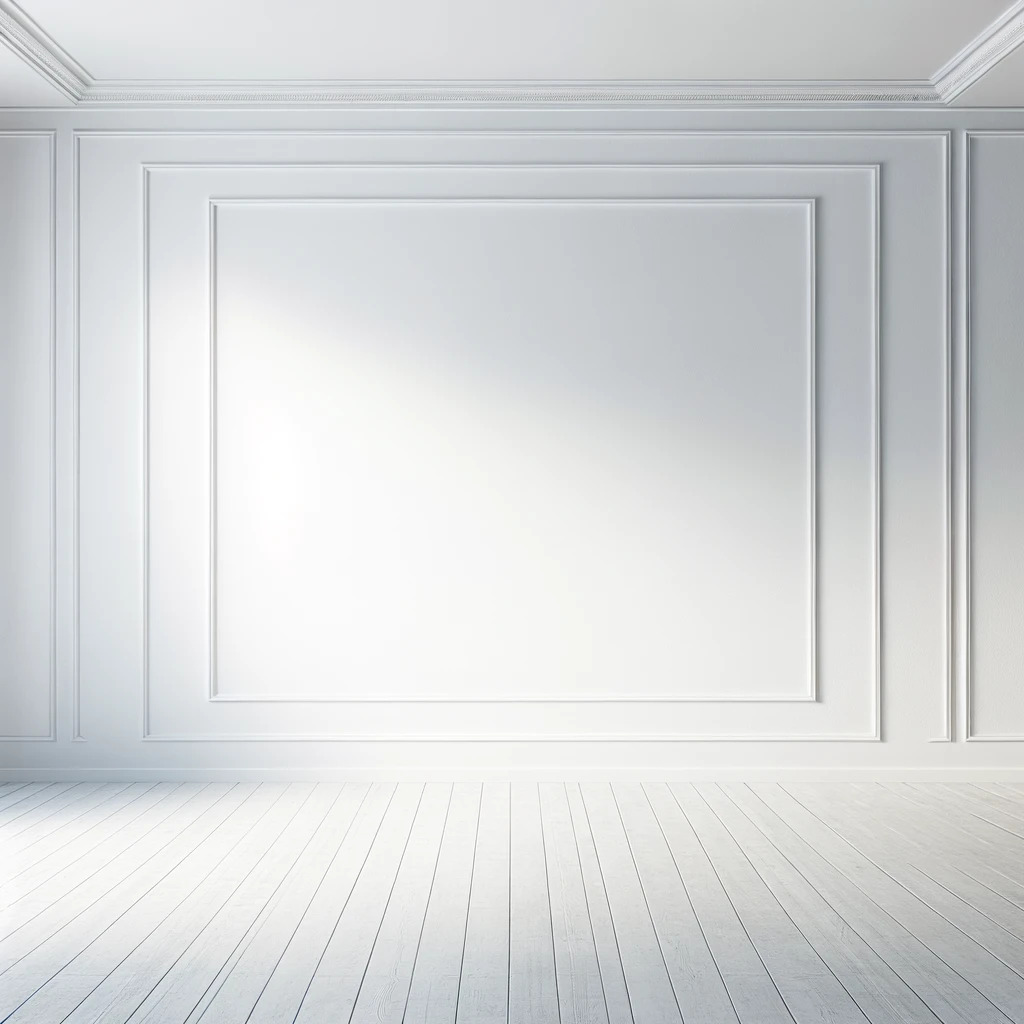}
    \label{fig:intro:simple}
  }
\subfloat[\textit{``a colorful park with a crowd''}]{
    \includegraphics[width=0.47\linewidth]{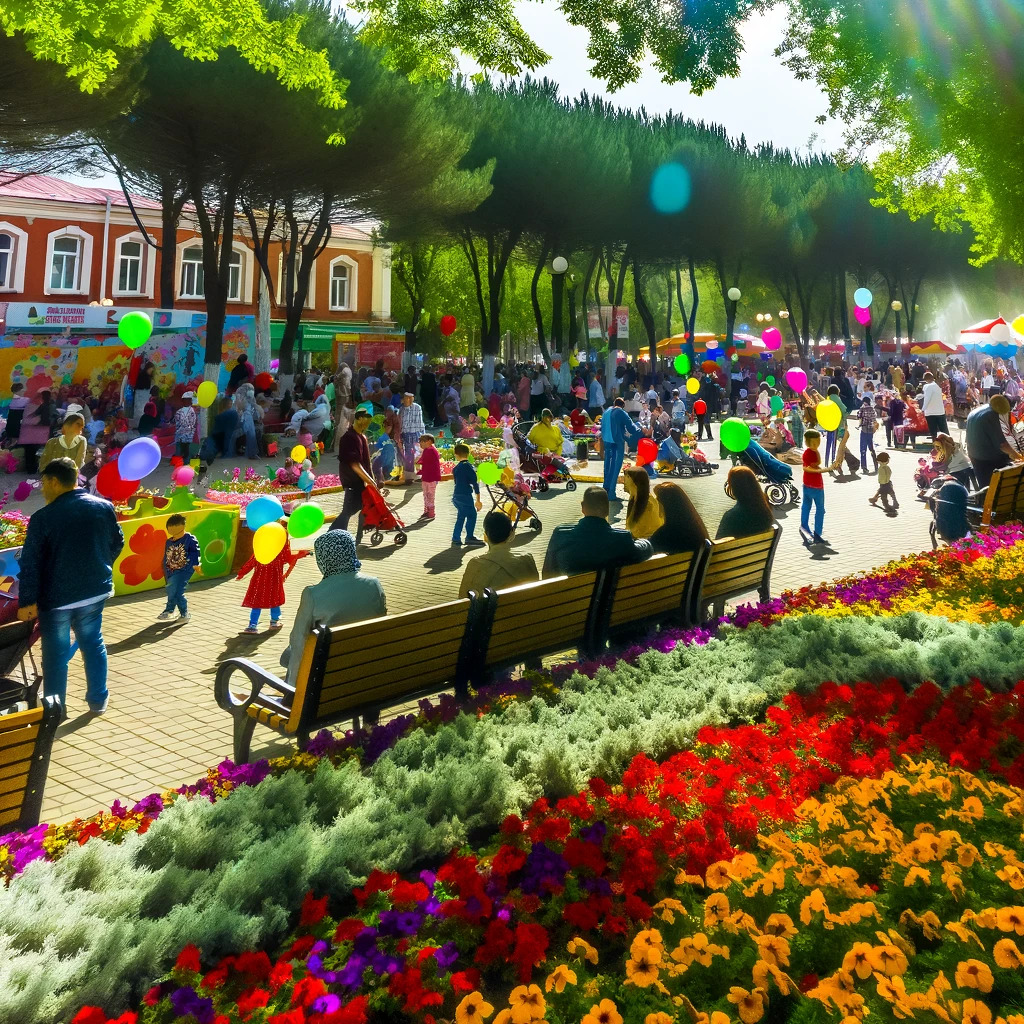}
     \label{fig:intro:complex}
  }
\Caption{Example generative images with different visual complexity.}
{With current diffusion models, the two images are generated with the same denoising steps and computational cost. However, intuitively, the simpler \subref{fig:intro:simple} could have been generated with less computation than \subref{fig:intro:complex}. 
This insight motivates us to develop \methodName, an efficiency-optimized guidance for balancing quality vs. computation trade-offs. \methodName tailors the denoising process to align with the given prompt before generation starts.}
\label{fig:intro}
\Description{}
\end{figure}

A significant body of recent literature has attempted to improve the efficiency of denoising diffusion models, such as reducing the number of denoising steps via neural network simplifications \cite{li2024snapfusion,li2023autodiffusion} and model distillation \cite{sauer2024fast, salimans2022progressive, meng2023distillation, liu2023instaflow}.
However, such one-size-fits-all approaches inevitably lead to images generated with inconsistent quality depending on the desired content reflected in the text prompts. 
\nothing{
Meanwhile, the entire denoising process is pre-determined by the number of total denoising steps before generation, prohibiting run-time measurement and early termination. \qisun{@kenny: help me with a reference.}
}
\nothing{
\warning{mention about skip steps, and therefore diffusion step optimizations cannot be implemented as observe-on-the-go.}
\qinchan{Essentially, the denoising process of the diffusion models is to solve the Ordinary Differentiable Equation(ODE) or Stochastic Differentiable Equation (SDE) to get X(0) which is the generated image in principle with a known X(T) which is the initial Gaussian Noise. And dX/dt at each timestep t is commonly estimated by UNET or ViT. Therefore, we must run the whole process to timestep 0 to get X(0), which is by definition the generated images, no matter how timesteps from T to 0 are skipped or scheduled in different schedulers or techniques. (evenly or unevenly, linearly or not, number of skipped steps, etc).}
}
\note{
What our method has to offer, sales pitch for concrete benefits, not technical details.
Imagine we are doing a TV advertisement here.
}
For example, consider the two prompts: 1) ``a white and empty wall'' vs. 2) ``a colorful park with a crowd'' as shown in \Cref{fig:intro}. 
Intuitively, the two prompts may require different number of inference steps to generate images of high quality. 
With this inspiration, we present \methodName, a perceptually-aware guidance model for text-to-image diffusion models.
\note{
Our main idea, giving people a take home message and (if possible) see how clever we are.
}
Specifically, for a given text prompt, \methodName estimates the correlation between various visual quality metrics and number of diffusion inference steps as a time series function.
The estimation then suggests the optimal number of denoising steps with a gradient-based analysis.
That is, we determine the plateau point where additional denoising steps will benefit only marginally in terms of visual perception.
\note{
Our algorithms and methods to show technical contributions and that our solutions are not trivial.
}
To achieve this goal, we first created a large-scale generative dataset using more than 18,000 text prompts $\times$ 12 timesteps.
By leveraging this data, we can measure perceptual similarity at various scales, ranging from per-pixel errors to spatial layout and semantics.
From this perceptual assessment of generative images and the Markov Chain nature of diffusion models, we developed a long short-term memory (LSTM) neural network \cite{graves2013speech} that predicts the relationship between quality enhancement and diffusion steps.

\note{
Results, applications, and extra benefits.
}
We conduct a series of objective analyses and subjective user studies. 
Our findings demonstrate \methodName's effectiveness in significantly enhancing the computational efficiency of diffusion models, measured as perceptual gain per diffusion step.
Moreover, this enhancement is achieved without compromising subjectively perceived quality. 
We hope this work will spark a series of future initiatives on human-perception-centered approaches for  computation- and energy-friendly generative models\footnote{We will release the code and data upon acceptance.}.
\section{Related Work \underRevision}
\label{sec:related}


\subsection{Optimizing Generative Diffusion Models}
Diffusion models have recently exploded in popularity due to their high performance on tasks such as image and video generation, audio generation, and 3D shape generation \cite{ho2020denoising,ramesh2021zero}.
Latent diffusion models \cite{rombach2022high} have significantly improved training and inference efficiency, but still require a high number of forward denoising neural network evaluations to produce high-quality results. 
To tackle this problem, an extensive body of literature has been proposed to optimize and accelerate diffusion models from different perspectives.
For example, optimizing the sampling strategy may enable more efficient denoising computation \cite{li2024snapfusion,chen2023speed,li2023autodiffusion}, such as timestep integration \cite{nichol2021improved} or conditioning on the denoising \cite{preechakul2022diffusion}.
Additionally, by approximating the direct mapping from the initial noise to generated images \cite{luo2023latent,song2023consistency}, denoising can be executed with a reduced number of steps. 
%
%
%
After the models are pre-trained, distilling them to student models can generate specific images with fewer steps \cite{sauer2024fast,salimans2022progressive,meng2023distillation,liu2023instaflow}.
Optimized solvers for the denoising step can efficiently reduce the computation to avoid re-training or fine-tuning \cite{song2020denoising,lu2022dpm,liu2022pseudo,karras2022elucidating}.

While reduced computation often results in degraded image quality, existing approaches typically prioritize computational models over human perception. For instance, as shown in \Cref{fig:intro}, different text prompts could be allocated with varying levels of computation to achieve similar visual quality from a human perspective.
To our knowledge, the perceptual correlation between prompt-indicated visual content complexity and computational cost remains unknown and is not yet integrated into the diffusion model workflow.
In this research, our \methodName introduces the first complementary human perception-aware approach, guiding diffusion computation toward the most efficient ``perceived quality gain per step'' tailored to the prompt.

\subsection{Perceptually-Guided Computer Graphics}
A wealth of literature in the graphics community has studied applications of human perceptual data to optimize visually-based algorithms.
For example, several image quality metrics have been proposed which generally agree with experimental data, such as SSIM \cite{wang2004image} or PSNR, or neural network-based metrics like LPIPS \cite{zhang2018unreasonable}.
Metrics which are based on psychophysical models are more highly correlated with human responses, and can model temporal, high luminance, and foveated artifacts \cite{daly1992visible,mantiuk2011hdr,mantiuk2021fovvideovdp}.
Recently, metrics such as DreamSim \cite{fu2023learning} have been proposed to determine higher-order image differences, such as layout or semantics.
Several techniques are built with human perception in mind, such as for optimizing traditional rendering pipelines \cite{patney2016towards,jindal2021perceptual} or for improving the performance of generative models \cite{czolbe2020loss,johnson2016perceptual}.
However, there has been little investigation into understanding and optimizing generative content from the perspective of perceptual guidance.

\section{Method\feedbackNeeded}
\label{sec:method}
 

\begin{figure*}
  \centering
  \subfloat[our pipeline]{
    \label{fig:overview:pipeline}
    \includegraphics[width=.47\linewidth]{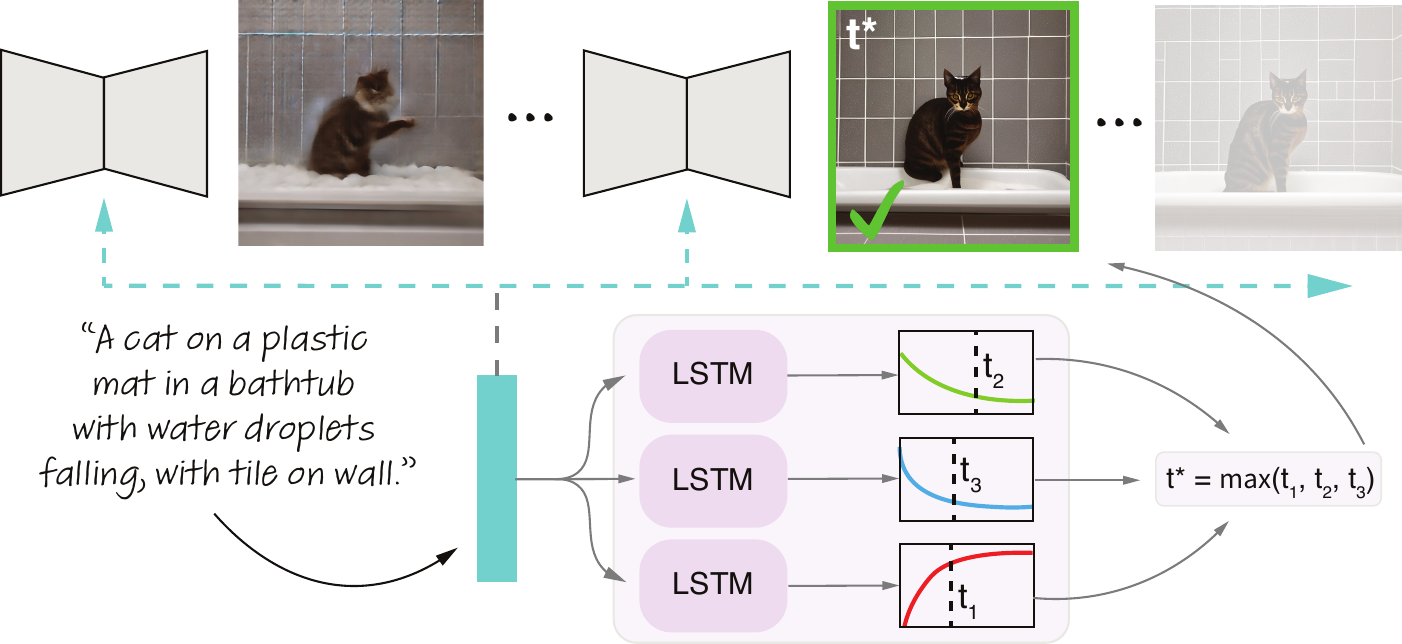}
  }
  \subfloat[example with our model guidance]{
    \label{fig:overview:example}
    \includegraphics[width=.51\linewidth]{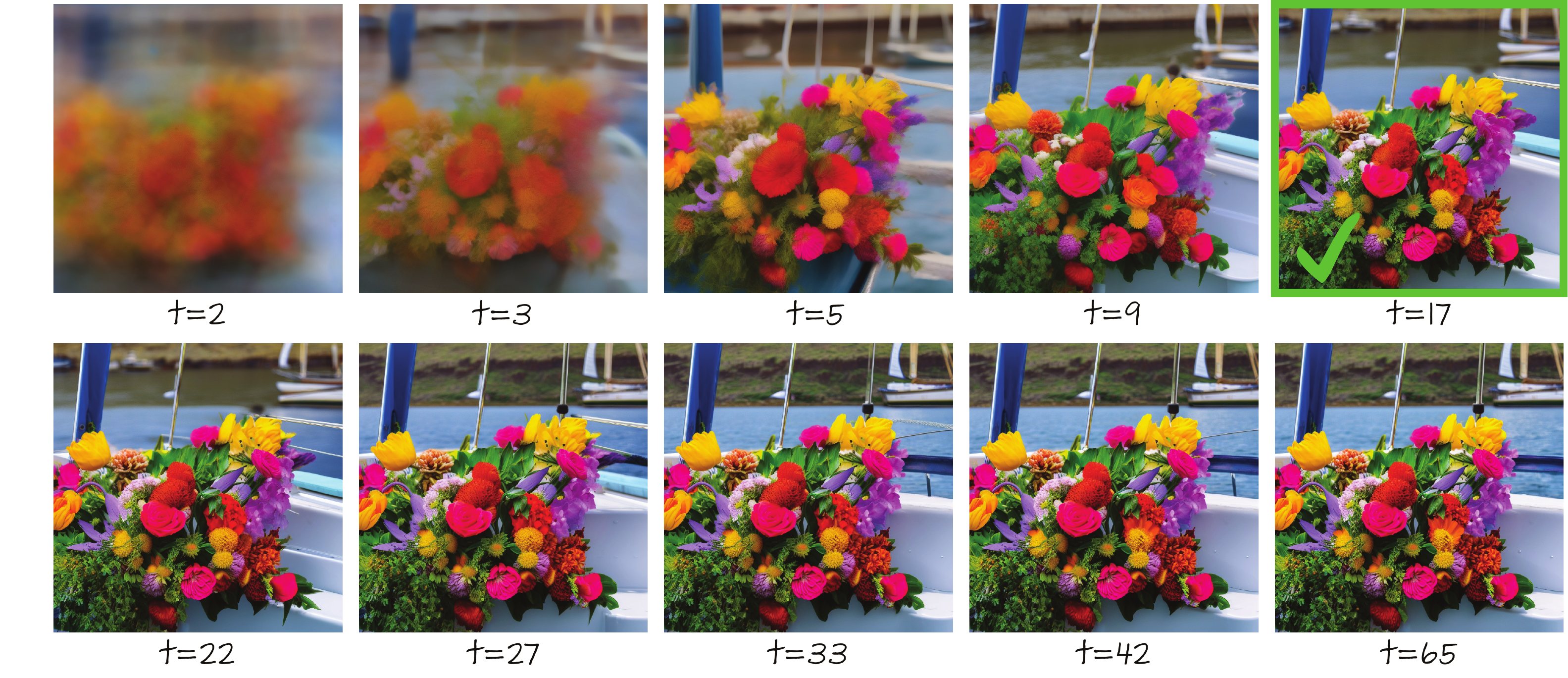}
  }  

  \Caption{
      Our \methodName pipeline.
  }{
    \subref{fig:overview:pipeline} Given an input prompt, we predict three time series of perceptual quality metrics of the generated images at different timesteps (the three colored curves). Each one of them represents a given perceptual scale.
    The model determines the optimal timestep, $\step^*$, which is the max plateau point of the three metrics, described in \cref{sec:method:metrics}.
    The pre-trained diffusion model performs $\step^*$ number of denoising steps, rather than continuing the forward process, which would only yield little image quality improvement as predicted by our model.
    \subref{fig:overview:example} 
    We include an additional example of the forward process and the selection made by our model.
  }
  \label{fig:overview}
  \Description{}
\end{figure*}

We develop \methodName to suggest the optimal number of inference, or denoising, steps based on predicted perceptual quality.
Toward this aim, and as visualized in \Cref{fig:overview}, we first created a dataset of images generated with different numbers of denoising steps (\Cref{sec:method:data}). 
To comprehensively measure the perceptual quality of this large-scale synthetic dataset, we leveraged perceptual metrics at various levels of detail, including their pixel-wise error, layout, and overall semantic similarity with respect to "reference" images generated with a large number of inference steps (\Cref{sec:method:metrics}).
With the jointly labelled data of text prompt, perceptual scores, and denoising steps, we train an LSTM-based \methodName model to predict metric scores against denoising steps as a time series function, given input prompts (\Cref{sec:method:text2score}).
Lastly, we employ \methodName to estimate the least required number of denoising steps which produces high-quality images using plateau point estimation (\Cref{sec:method:score2step}).

\subsection{Synthetic Dataset Generation} 
\label{sec:method:data}
We first generated an image dataset by running the forward pass of a diffusion model at different number of denoising steps in order to understand the change in perceptual quality throughout timesteps. 
To this end, we sample a large set of text prompts and denoising steps to be used to generate images with a diffusion model.

We used the popular pre-trained Stable Diffusion 2 model (SD)\footnote{https://github.com/Stability-AI/stablediffusion}, a variant of Stable Diffusion \cite{rombach2022high}, as a drop-in to demonstrate \methodName.
Stable Diffusion 2 generates high-quality images despite faster inference time compared to larger models, like Diffusion XL.
Note that \methodName may be applied to other diffusion model architectures without loss of generality. 

\paragraph{Creating and sampling prompts.}
To obtain prompts that cover a wide range of semantics from real-world scenarios, we use an existing image-caption pair dataset, COCO \cite{lin2014microsoft}. 
However, the dataset contains more than 330,000 images, making it computationally infeasible to generate a corresponding image for each of the captions as a prompt using diffusion models.
Therefore, we systematically sample the most representative captions using the Cosine dissimilarity score $\CLIPsimilarity$ in the CLIP-encoded space \cite{radford2021learning}, where higher $\CLIPsimilarity$ indicates more similar text pairs.
%
%
That is, to construct the prompt set $\selectedPromptSet$, we keep the prompts so that their pair-wise CLIP similarity is lower than a threshold
%
\begin{align}
\similarity(\prompt, {\prompt}^\prime) < \thresholdClip,\  \forall \prompt, {\prompt}^\prime\in\selectedPromptSet.
%
\label{eqn:CLIPsimilarity}
\end{align}
We experimentally set the value of $\thresholdClip$ to $.75$, and acquired 18,384 prompts to compose $\selectedPromptSet$.
The sampled text prompts are then used to generate images at different denoising steps using SD. 

\paragraph{Sampling denoising timesteps.}
In most existing diffusion models, the computation could be reduced by skipping denoising timesteps with the cost of reduced quality. 
Therefore, the main aim of \methodName is to suggest the most efficient trade-off between perceptual quality and computation cost based on a specific text prompt. 
%
Toward this aim, for all text prompts in $\selectedPromptSet$ we generate images at various denoising steps to make a systematic measurement of the change in image quality at each step, as shown in \Cref{fig:overview:example}.
Because it is computationally infeasible to perform image generation for all possible timesteps, we consider a power law as guidance to sample the diffusion timestep set $\stepSet\coloneqq\{\step_1, ..., \step_\stepSetSize\}$.
Specifically, we select $\step_i=1+2^{i-1}$, and additionally include a boundary timestep $\{t_1=1\}$ and empirically sample intermediate conditions, $\{22, 27, 42\}$, to further increase the sampling density. 
We sample this range up to 129 ($i=8$), which is above the commonly accepted sufficient range of 100 inference steps for Stable Diffusion \cite{nguyen2024inference}. 
This sampling strategy results in $\stepSetSize=12$ time steps in total.

\paragraph{Scheduler and generation.} 
\warning{as diffusion has to be skipping connect, we cannot observe-on-the-goal.} \qinchan{as stated in the Introduction.}
All the experiments in our work use the widely used Euler Scheduler \cite{karras2022elucidating}. 
We make this choice due to the scheduler's more consistent and cohesive evaluation results because of its fixed start (1) and end (1000) timesteps. 
%
We have 4 fixed random seeds as the generator of the initial noise and PyTorch manual random seeds that control the stochastic denoising process. \qisun{I do not understand this sentence}
Using the sampled conditions, we generate an image $\image_{\prompt,\step}$ from each of the selected prompts in $\selectedPromptSet$ at each representative time step in $\stepSet$ and $4$ random seeds using SD. Overall, the dataset contains $18,384\ ($\prompt$)\, \times\, 12\ ($\stepSetSize$)\ \times \ 4 (\text{seeds})=882,432$ images of resolution $768\times 768$.
We reserved the images from 90\% of randomly sampled prompts as the training set (\Cref{sec:method:text2score}) and the remaining 10\% for evaluation (\Cref{sec:result}).

\subsection{Multi-Scale Perceptual Metrics}
\label{sec:method:metrics}

We measure the perceptual quality throughout the synthetic dataset of an image $\image$ corresponding to prompt $\prompt$ and timestep $\step$, denoted as $\{\image_{\prompt,\step}\}$.
The denoising process over $\step$ iteratively removes noise to generate the final image. 
Each step jointly enhances local details and/or global image-text alignment \cite{yang2023diffusion}. 
Therefore, image quality should be measured at various perceptual scales in terms of level of detail and semantics. 
To this end, we propose to leverage three perceptual metrics to compare images generated at each step.
First, we measure the most detailed pixel-level quality using Laplacian signal-to-noise ratio (\SNRLaplacian) \cite{kellman2005image}. 
Then, we leverage the recent learning-based layout-aware DreamSim metric (\dreamsim) \cite{fu2023learning} to measure mid-level content similarity. 
Lastly, we measure the high-level semantic alignment using the Cosine similarity in CLIP-encoded latent space (\clip) \cite{radford2021learning}.
Both \dreamsim and \clip metrics compare each image $\image_{\prompt,\step}$ against the target image ($\image_{\prompt,\step_\stepSetSize}, \step_\stepSetSize=129$).

\qinchan{do we have to mention that we used the divide by 3 normalization techniques to get this metric to [0,1]?}
\paragraph{Pixel-level: Laplacian signal-to-noise ratio \textbf{(\SNRLaplacian$\in[0,1]\downarrow$)}} 
An insufficient number of forward denoising steps may generate noisy and distorted images \cite{yang2023diffusion}. 
Therefore, we first measure each image's pixel-level quality. 
The signal-to-noise ratio (SNR) is a widely adopted metric for image quality \cite{kellman2005image} and deep metric learning \cite{yuan2019signal}. 
To our aim, we employ the SNR as a no-reference \SNRLaplacian metric, which assesses the SNR by comparing the original image to its Laplacian-filtered version, thereby evaluating its sharpness \cite{paris2011local}.
\begin{align}
    \text{\SNRLaplacian}\left(\prompt,\step\right) \coloneqq \text{SNR}\left(\image_{\prompt,\step}, \GaussianKernal\circledast\image_{\prompt,\step}\right),
    \label{eqn:SNRfinal}
\end{align}
where $\GaussianKernal$ and $\circledast$ are the Gaussian kernel ($\sigma=1$) and convolution operator, respectively.
%
We developed this measurement instead of using reference-based image quality metrics (e.g., PSNR) against the target image $\image_{\prompt,\step_\stepSetSize}$ due to the stochastic and random nature of diffusion model-based generation, where pixel-level values can significantly vary based on the added random noise.

\paragraph{Mid-level: DreamSim \textbf{(\dreamsim$\in[0,1]\downarrow$)}}
The DreamSim metric was recently proposed by \citet{fu2023learning}.
It measures mid-level perceptual similarity between two images, considering non-pixel factors such as spatial layouts. 
We compute the \dreamsim metric as
\begin{align}
\text{\dreamsim}(\prompt,\step) \coloneqq \dreamsimscore(\image_{\prompt,\step}, \image_{\prompt,\step_\stepSetSize}),
\end{align}
where $\dreamsimscore$ indicates the DreamSim distance.

\paragraph{Semantic-level: image caption encoded by CLIP \textbf{(\clip$\in[0,1]\uparrow$)}}
The CLIP model \cite{radford2021learning} maps text and image embeddings into the same latent space, and has been used for image captioning \cite{lin2014microsoft,barraco2022unreasonable} and other text-conditional generation tasks, such as image generation\cite{wang2022clip,rombach2022high}, and 3D mesh generation \cite{mohammad2022clip}.
On the highest level, a poorly generated image may semantically mis-align with an input prompt \cite{yang2023diffusion}.
Therefore, determining the captions between a pair of images may reflect their contextual and semantic distances. 
With this observation in mind, we generate the CLIP-encoded caption for each image, and compare their distance in the encoding space,
\begin{align}
\text{\clip}\left(\prompt,\step\right) \coloneqq \cos \left(\CLIPembedding{\CLIPcaptioning}\left(\image_{\prompt,\step}\right),\CLIPembedding{\CLIPcaptioning}(\image_{\prompt,\stepSetSize})\right),
\label{eqn:metric:clip}
\end{align}
where $\CLIPcaptioning$ is CLIP captioning, and $\CLIPembedding{\CLIPcaptioning}$ is the latent embedding.

\subsection{Learning to Predict Perceptual Metrics from Text}
\label{sec:method:text2score}
The multi-scale perceptual analysis of each image $\image_{\prompt,\step}$ creates a large-scale dataset of prompt - timestep pairs which map to the three metrics scores computed on images generated at the corresponding timestep by averaging all scores over the four random seeds,
\begin{align}
\left(\prompt,\step\right)\rightarrow\{\text{\SNRLaplacian}\left(\prompt,\step\right),\text{\dreamsim}\left(\prompt,\step\right),\text{\clip}\left(\prompt,\step\right)\}.
\end{align}
For each prompt $\prompt$, the per-time step data informs the perceptual quality at a given denoising timestep $\step$.
We then trained neural networks $\lstm_\metric$ to predict each of the above time series for novel text prompts. Here, $\metric\in\{\text{\SNRLaplacian},\text{\dreamsim},\text{\clip}\}$ are the individual perceptual metrics. 

\paragraph{Model architecture}
\begin{figure}
\centering
\includegraphics[width=\linewidth]{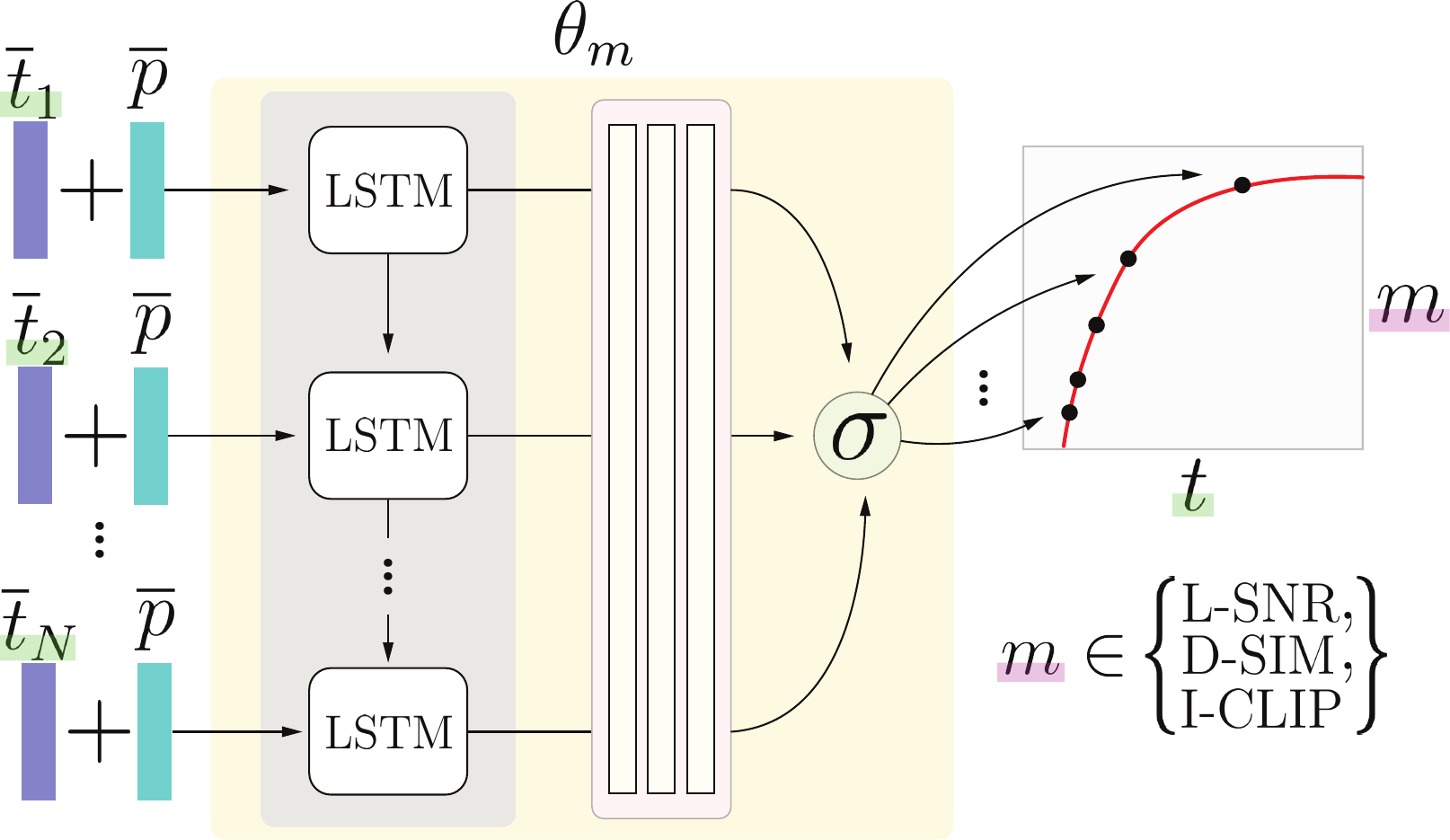}
\Caption{Model architecture.}
{
We visualize our architecture as an unrolled LSTM, $\lstm_\metric$, which takes as input positionally-embedded timestep, $\positionEmbedding{\step}_i$, and clip-embedded prompt, $\CLIPembedding{\prompt}$.
Outputs for each timestep are fed to fully-connected layers, and normalized with a sigmoid activation to produce scores, $m$, which are one of three perceptual metrics defined in \Cref{sec:method:metrics}.
}
\label{fig:model_architecture}
\Description{}
\end{figure}
To simulate the change in image quality with respect to the number of denoising steps, the model architecture should reflect the 
locally dependent chain nature of the denoising process of diffusion models.
With this insight, we choose to use a bidirectional LSTM (BiLSTM) \cite{graves2005framewise} given the robustness and accuracy observed for various similar time series prediction tasks \cite{siami2019performance}.
Our model architecture is visualized in \Cref{fig:model_architecture}.
To transfer the stochastic chain process to a recurrent form for the LSTM, the CLIP-embedded text prompt $\CLIPembedding{\prompt}$ and position-embedded (as in \citet{kenton2019bert}) timestep $\positionEmbedding{\step}$ vectors are combined as input to predict metric scores at timestep $\step$,
\begin{align}
    \metric_\step(\prompt) = \lstm_\metric(\CLIPembedding{\prompt}+\positionEmbedding{\step}).
\end{align}
The implementation details of our model are discussed below. \nothing{\qinchan{\cite{souza2022bert}, \cite{cheragui2023exploring}, \cite{oliveira2020optic}, and \cite{mahadevaswamy2023sentiment} place a feed-forward network (fully connected layer, or MLP) after LSTM/Transformer to fine-tune or train their models for sentiment analysis and POS tagging.}}

\paragraph{Model implementation details}
As shown in \Cref{fig:model_architecture} and similar to prior literature \cite{souza2022bert,cheragui2023exploring,oliveira2020optic,mahadevaswamy2023sentiment}, we implement $\lstm_\metric$ as a simple BiLSTM, followed by a feed-forward fully connected network (MLP) and Sigmoid activation.
Specifically, for the LSTM, we use a two-layer BiLSTM with hidden features of size 512. The three-layer MLP transforms the LSTM-returned hidden scores of size 1024 (due to the bidirectional design) to 128 and then 1.
We train the model with our held-out 90\% (\Cref{sec:method:data}) training images from 16,545 prompts 
 with L2 loss, a batch size of 32, and a learning rate of 1e-4. During training, we used the widely adopted DropOut \cite{hinton2012improving} technique on the output of the bidirectional LSTM.
To enhance the fitting ability of the model, we linearly interpolate the scores from sampled steps in $\stepSet$ to get pseudo-ground-truth scores for all other steps. Finally, we train our model for \SNRLaplacian and \dreamsim with 25 epochs and for \clip with 50 epochs.

\subsection{Predicting Denoising Steps from Perceptual Metrics}
\label{sec:method:score2step}
\begin{figure}
\centering
\includegraphics[width=\linewidth]{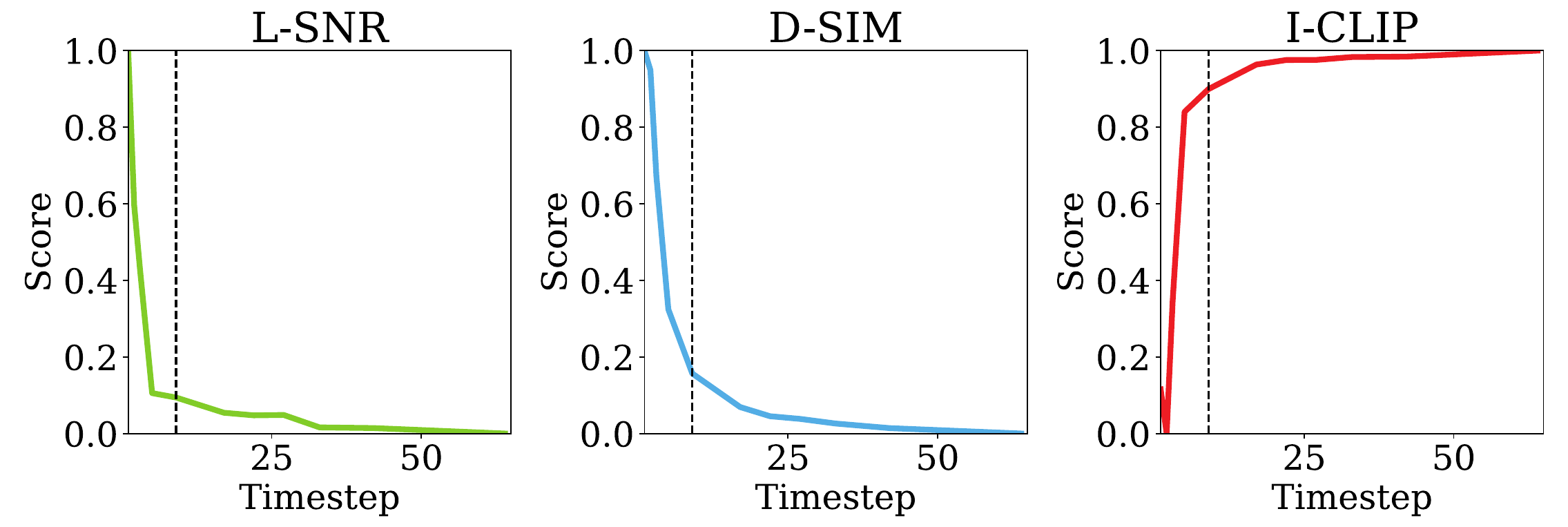}
\Caption{Example results of predicting denoising steps before generation.
}
{By leveraging predicted perceptual metrics with regard to timesteps as a time series, we predict the ``plateau points'' for each. Their max (\Cref{eqn:optimization}) suggests the most efficient timesteps for diffusion models, before the generation starts.
}
\label{fig:method:cut}
\Description{}
\end{figure}
\begin{figure}
\centering
\subfloat[insufficient \SNRLaplacian]{
    \includegraphics[width=0.32\linewidth]{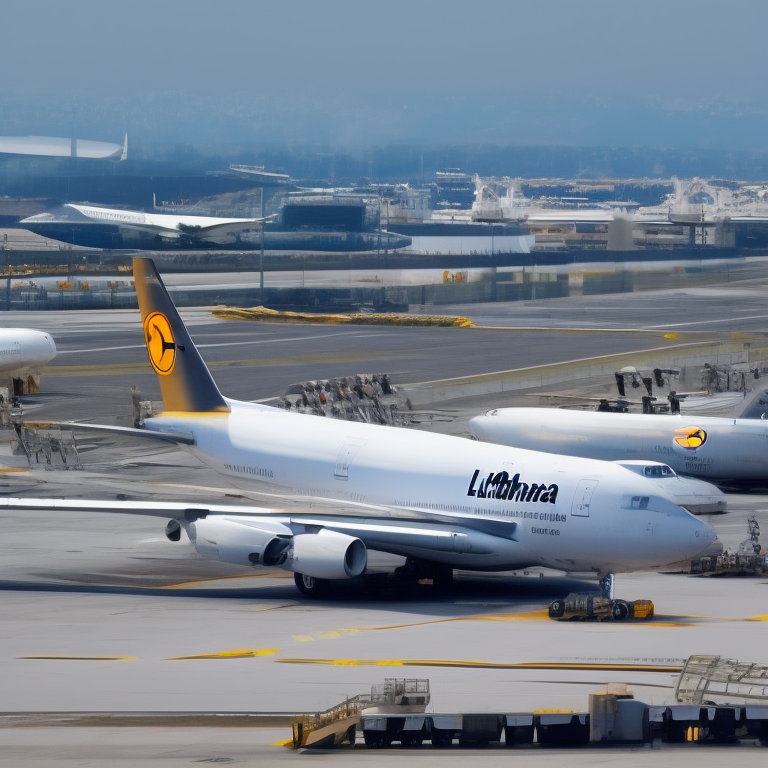}
    \label{fig:eval:effective:l-snr}
  }
\subfloat[insufficient \dreamsim]{
    \includegraphics[width=0.32\linewidth]{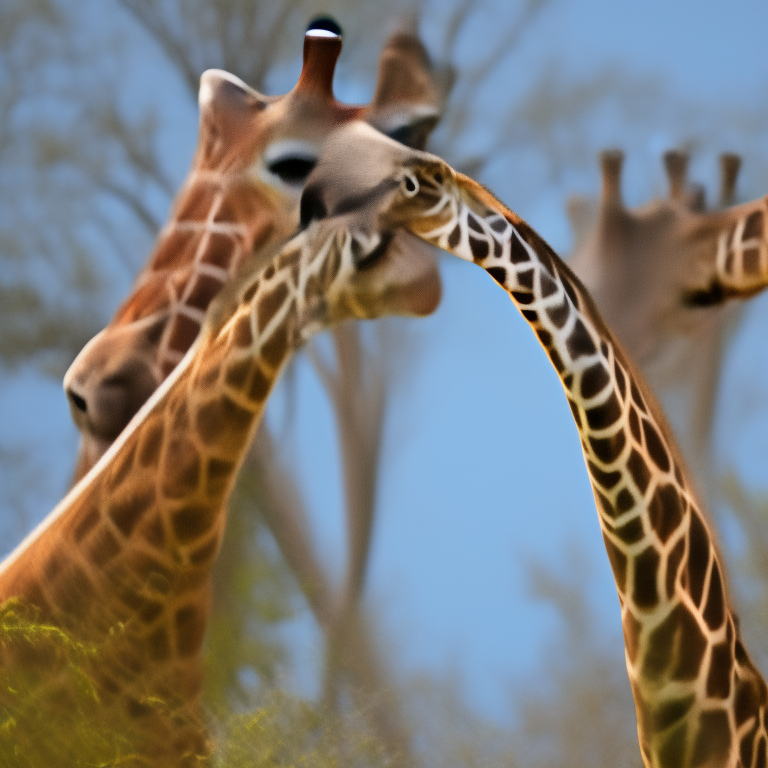}
    \label{fig:eval:effective:l-dsim}
  }
\subfloat[insufficient \clip]{
    \includegraphics[width=0.32\linewidth]{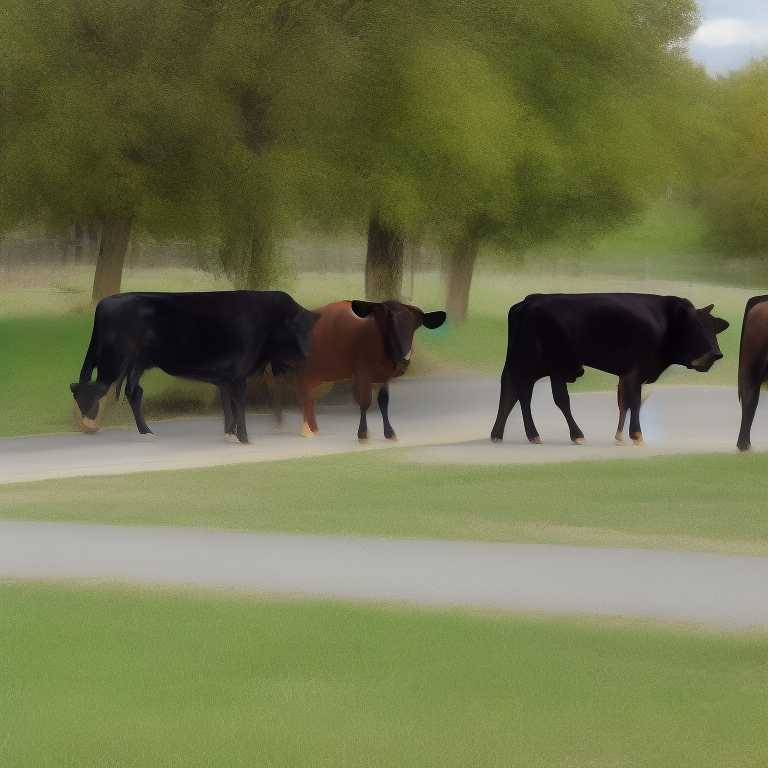}
    \label{fig:eval:effective:l-clip}
  }  

\vspace{-1em}
\subfloat[\methodName \SNRLaplacian]{
    \includegraphics[width=0.32\linewidth]{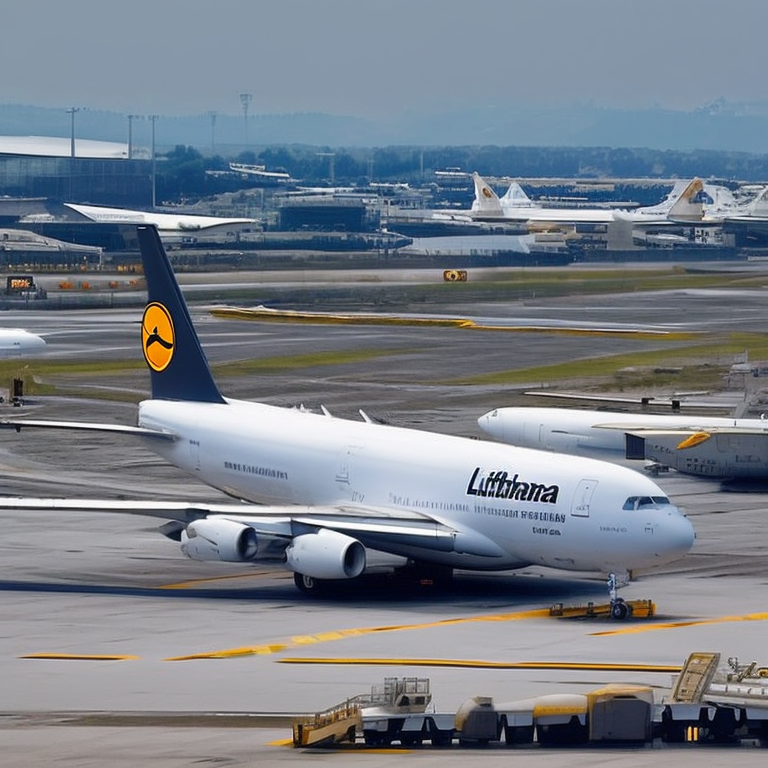}
    \label{fig:eval:effective:h-snr}
  }
\subfloat[\methodName \dreamsim]{
    \includegraphics[width=0.32\linewidth]{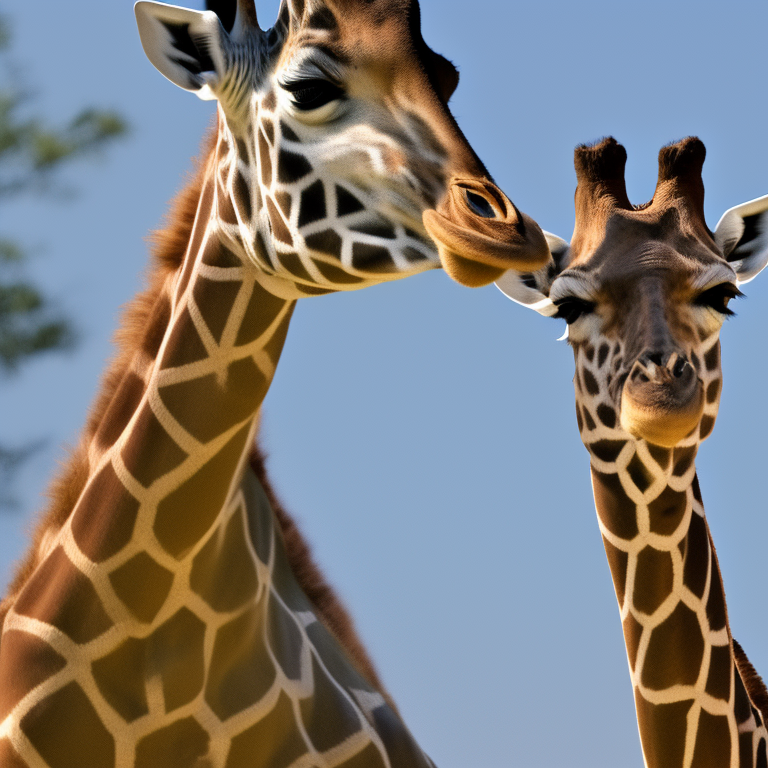}
    \label{fig:eval:effective:h-dsim}
  }
\subfloat[\methodName \clip]{
    \includegraphics[width=0.32\linewidth]{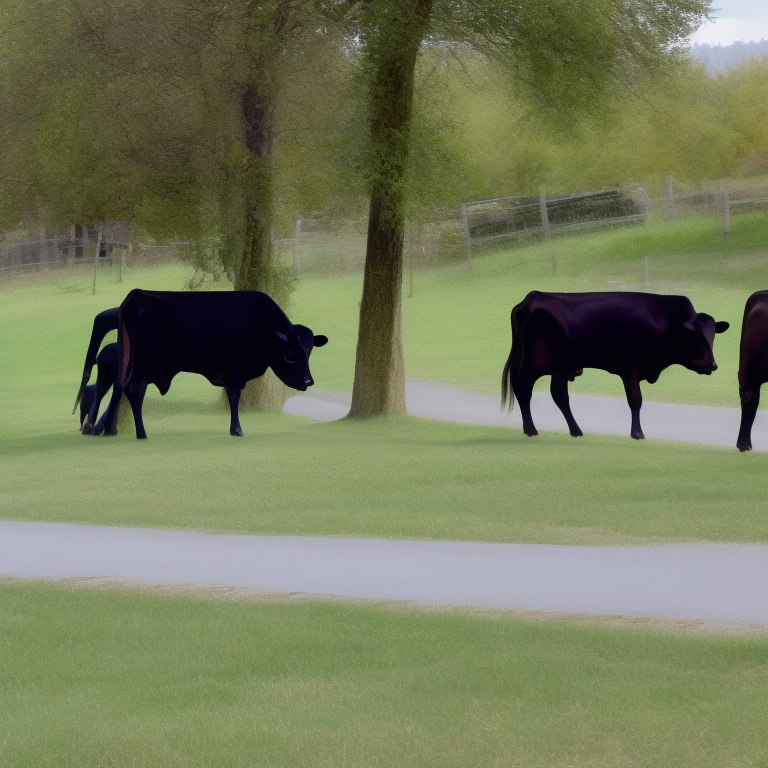}
    \label{fig:eval:effective:h-clip}
  }    
\Caption{Sufficiency of \methodName-suggested denoising steps.}
{%
The first/second row visualizes example images with step numbers below/from our model suggestion. Each column compares the effectiveness of corresponding perceptual metrics.
\subref{fig:eval:effective:l-snr} vs. \subref{fig:eval:effective:h-snr}: insufficient denoising steps caused blurry image, reflected as low \SNRLaplacian values.
\subref{fig:eval:effective:l-dsim} vs. \subref{fig:eval:effective:h-dsim}: insufficient denoising steps caused structural and geometric distortion, reflected as low \dreamsim values.
\subref{fig:eval:effective:l-clip} vs. \subref{fig:eval:effective:h-clip}: images generated from text prompt ``\textit{black} cows following on another across the street[...]''. Insufficient denoising introduces brown cows that are semantically undesired from the prompt. 
}
\label{fig:eval:effective}
\Description{}
\end{figure}

For a new prompt $\prompt$, we leverage the trained LSTM models $\lstm_\metric$ to predict the time series $\metric_\step$ for each $\metric\in\{\text{\SNRLaplacian},\text{\dreamsim},\text{\clip}\}$.
Our goal is to suggest the total denoising timestep of the diffusion models before generation. 
As shown in \Cref{fig:method:cut}, we intuitively find the ``turning'' timestep $\stepOptimal$ where all metrics plateau. That is, additional inference steps ($t>\stepOptimal$) will only improve users' perceived image quality marginally at all scales, indicating reduced computational efficiency.
Note that the perceptual metrics typically exhibit a monotonic relationship with regard to number of denoising timesteps, i.e., larger timesteps do not reduce the perceived quality. 
Inspired by prior literature on turning point detection for psychological time series data \cite{engbert2006microsaccades}, we employ a statistical approach to determine optimal time step,
\begin{align}
\begin{split}
    \stepOptimal(\prompt) &\coloneqq \max{\stepOptimal_\metric(\prompt)},\ \metric\in\{\text{\SNRLaplacian},\text{\dreamsim},\text{\clip}\}\\
    \stepOptimal_\metric(\prompt) &\coloneqq \max_\step\ \text{s.t.}\ \metric_\step(\prompt) \geq \mean\left(\{\metric_\step(\prompt)\}_{\step=\step_1}^{\step_\stepSetSize}\right) + \weightPlateau_\metric \std\left(\{\metric_\step(\prompt)\}_{\step=\step_1}^{\step_\stepSetSize}\right).
    \label{eqn:optimization}
\end{split}    
\end{align}
Here, $\mean(\cdot)/\std(\cdot)$ indicates the median and standard deviation of the time series data. 
Similar to \citet{engbert2006microsaccades}, we determine the weights $\weightPlateau_\metric$ ($\weightPlateau_{\text{\SNRLaplacian}}=0.3, \weightPlateau_{\text{\dreamsim}}=0.2, \weightPlateau_{\text{\clip}}=0.5$) of the three metrics using the efficiency measurement as detailed in \Cref{subsec:efficiency}.
The effectiveness of the suggested step numbers on each perceptual metrics is visualized in \Cref{fig:eval:effective}.

\nothing{
We measured the plateau in image quality by first detecting when the sharp decrease in the rate of improvement of scores occurred. For time t, and corresponding $score_t$:
\begin{align}
final\_scores = \sigmoid(\mlp(\scores))
\label{eqn:score_pred}
\end{align}
Here we used the 0.04 threshold for the SNR\_Laplacian score and the 0.3 threshold for the DreamSim score. However, first-order derivatives are vulnerable to perturbation, so we propose an additional numerical threshold that is used to detect whether the score is within a small enough range of the final score; specifically, we use 0.03 for both SNR\_Laplacian and DreamSim scores. \kenny{How were these values (1.3 and 1.5) chosen?}
}
\section{Evaluation \feedbackGiven}
\label{sec:result}

\note{1. compare ours + deepcache vs. deepcache. 2. quality per memory}

\begin{figure}
\centering

\includegraphics[width=\linewidth]{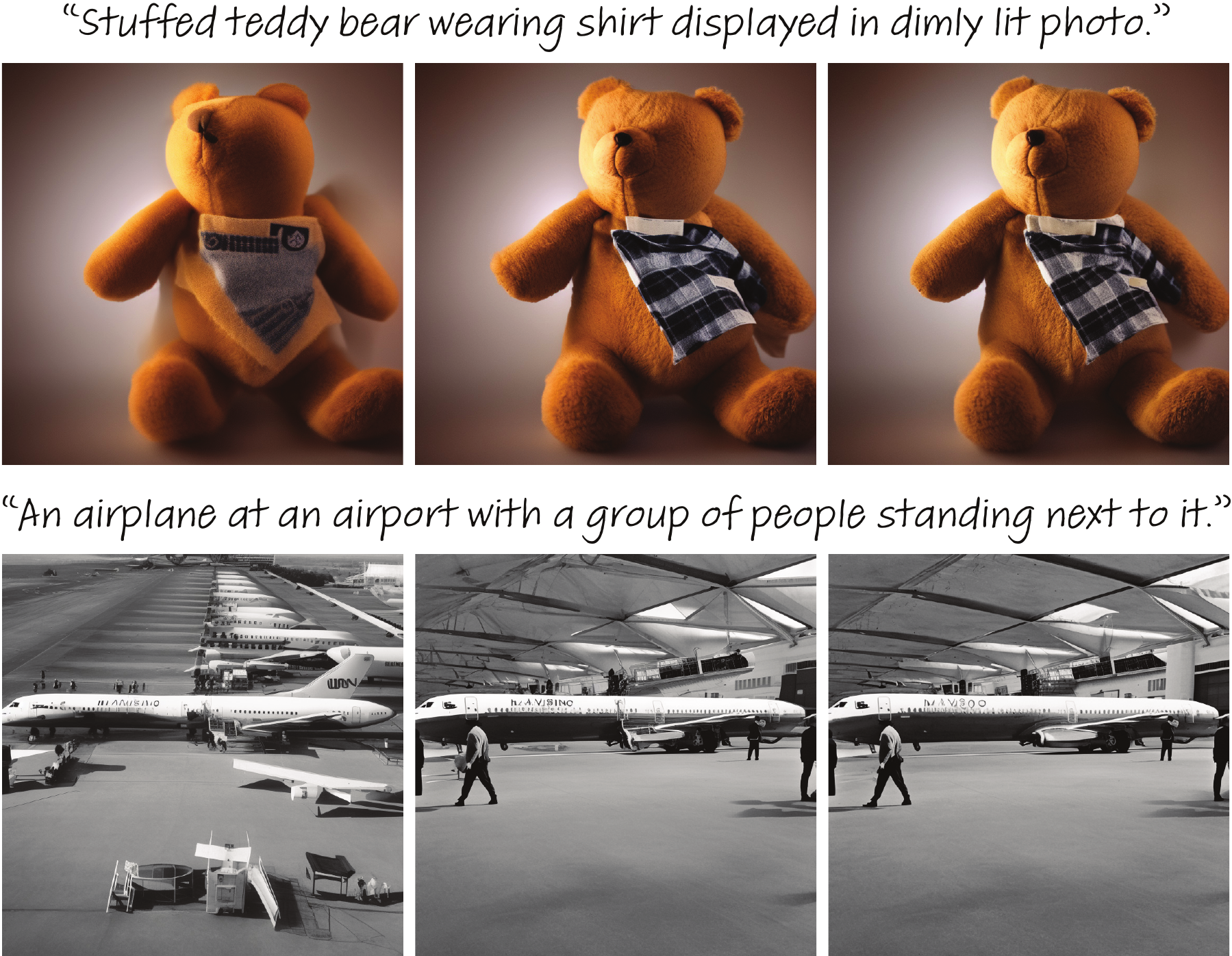}

\Caption{Qualitative results.}
{
We display two additional examples which show visible differences between subsequent timesteps.
The middle column in each row represents the images generated by the optimal timestep suggested by our model.
Their quality differences from the previous timestep (left) are large, but small from the next timestep (right). See supplementary material \nothing{\Cref{fig:tileResults} }for more cases.
}
\label{fig:qualitative2}
\Description{}
\end{figure}

Beyond the qualitative results demonstrated in \Cref{fig:qualitative2} and the supplementary material \nothing{\Cref{fig:tileResults}}, we first evaluate the model prediction accuracy as an ablation study in \Cref{subsec:accuracy}.
Then, we measure \methodName's effectiveness in enhancing computational efficiency as ``perceptual gain per diffusion step'' \Cref{subsec:efficiency}, as well as the generated content diversity \Cref{sec:eval:diversity}. Lastly, we conduct an user study for subjective measurement of overall generated image quality \Cref{sec:eval:study}.
To this aim, we study and compare three conditions:
\begin{itemize}[leftmargin=*]
    \item \condOurs: Our adaptive \methodName method;
    \item \condUniform: Instead of adaptive guidance, we construct a baseline which suggests a single timestep for all prompts. By calculating the average step numbers from \methodName throughout the evaluation set, this condition achieves, on average, the same computation cost to \condOurs; 
    \item \condRef: Within our sampled timesteps $\stepSet$, we select $\step=65$, which is close to the commonly and empirically suggested number of denoising steps. Higher step sizes will only  marginally enhance quality.
\end{itemize}
In the following experiment, we use the holdout test set of 1,839 prompts, which is 10\% of the entire image dataset.

\subsection{Ablation Study: Model Prediction Accuracy} 
\label{subsec:accuracy}
\begin{table}
\captionsetup{aboveskip=1.0ex}
    \caption{Ablation study of model MAE loss ($\downarrow$) for the perceptual metrics described in \cref{sec:method:metrics}.}
    \label{tab:modelpredictionchart}
    \resizebox{\columnwidth}{!}{
    \begin{tabular}{lrrr}
        \toprule
        Model & $|\Delta$\SNRLaplacian$|$  & $|\Delta$\dreamsim$|$  & $|\Delta$\clip$|$  \\
        \midrule
        \condOurs & \colorbox{green!30}{\textbf{.026}} & \colorbox{green!30}{\textbf{.037}} & \colorbox{green!30}{\textbf{.031}} \\ 
        \condOurs w/o t-encoding $\positionEmbedding{\step}$ & .059 & .239 & .134 \\
        MLP+Tanh replacing LSTM & .068 & .241 & .137\\
        \bottomrule
    \end{tabular}
    }
\end{table}

We evaluate our design of the proposed text-to-metric time series LSTM model (\Cref{sec:method:text2score}) as an ablation study. 
We computed mean absolute error (MAE, lower is better) between the predicted and measured ground truth to measure prediction accuracy.
Specifically, we compare the MAE of \condOurs, and \condOurs without text embedding (to validate the choice of embedding), as well as replacing our LSTM model with MLP + Tanh activation (to validate the choice of using LSTM).
\Cref{tab:modelpredictionchart} shows the results of our ablation.
\condOurs exhibits significantly lower error among all alternative design choices. 
The results evidence the effectiveness of the text prompt encoding and LSTM neural network.


\subsection{Quality and Computational Efficiency}
\label{subsec:efficiency}
The ultimate goal of \methodName is to optimize the ``quality gain per watt of computation'' of diffusion models. Therefore, we evaluate its quality versus its computation efficiency against the aforementioned conditions. 

\paragraph{Quality-Computation Efficiency}
We measure quality-computation efficiency as the perceptual quality gain per TFLOPS computation in the GPU. 
That is, for each evaluation prompt $\prompt$ and metric $\metric$, we compute
\begin{align}
\label{eq:gain}
\frac{\metric_\step\left(\prompt\right)}{\flops}.
\end{align}
Here, $\flops$ denotes the overall computation cost approximated as the TFLOPs (FLoating-point OPerations) during model inference. 
Higher average values indicate greater efficiency in quality gain.
Beyond the three conditions \condOurs ($\flops\approx68$ T), \condUniform ($\flops\approx67$ T), and \condRef ($\flops\approx394$ T), we further compared with alternative approaches in the literature on efficient diffusion models. In particular, we compared with the original denoising diffusion implicit models (DDIM) \cite{song2020denoising}, DeepCache \cite{ma2023deepcache}, DistriFusion \cite{li2024distrifusion}, as well as adding ancestral sampling with Euler method steps (EulerAncestral).
For fair comparison in computing the efficiency metrics, these approaches were controlled to consume the TFLOPs between \condUniform and \condOurs, with their corresponding reference images consumed approximately identical to \condRef.

\Cref{tab:avgqualitygain} presents the statistical summary across all evaluation prompts. The upper half shows that  \condOurs and \condUniform achieve significantly higher efficiency than \condRef. 
Compared with alternative methods in the lower half, \condOurs also exhibits superior performance. 
These results evidence the effectiveness of our method in achieving more quality-computation-balanced optimization for image generation diffusion models.

\begin{table}
\captionsetup{aboveskip=1.0ex}
    \caption{Quality-computation efficiency ($\mathbf{\uparrow}$) per step mean and standard error for different metrics and conditions (e-2 scaled). As raw \SNRLaplacian and \dreamsim values are negatively correlated to the quality, we inverted them here for consistency.}
    \label{tab:avgqualitygain}
    \resizebox{\columnwidth}{!}{
    \begin{tabular}{lrrr}
        \toprule
        Model & (1-\SNRLaplacian) $ / \flops$  & (1-\dreamsim)$/\flops$ &  \clip$/\flops$ \\
        \midrule
        \condOurs & \colorbox{green!30}{$\mathbf{0.46}$}$\pm 1.3\text{e-}2$  & \colorbox{green!30}{$\mathbf{1.38}$}$\pm 1.1\text{e-}2$ & \colorbox{green!30}{$\mathbf{1.48}$}$\pm 1.2\text{e-}2$  \\ 
        \condUniform & $0.42\pm 2.0\text{e-}3$ & $1.28\pm 1.8\text{e-}3$  & $1.36\pm 1.7\text{e-}3$  \\
        \condRef & $0.25$ & $0.25$ & $0.25$ \\
        \midrule[0.5pt]
        DDIM \cite{song2020denoising} & $0.41\pm2.1\text{e-}3$ & $1.17\pm 2.5\text{e-}3$ & $1.32\pm 2.0\text{e-}3$ \\
        DeepCache \cite{ma2023deepcache} & $0.43\pm2.0\text{e-}3$ & $1.27\pm 1.8\text{e-}3$ & $1.35\pm 1.7\text{e-}3$ \\
        DistriFusion \cite{li2024distrifusion} & $0.44\pm2.1\text{e-}3$ & $0.87\pm 3.1\text{e-}3$ & $1.22\pm 2.6\text{e-}3$ \\
        EulerAncestral & $0.38\pm1.8\text{e-}3$ & $1.0\pm2.6\text{e-}3$ & $1.26\pm2.4\text{e-}3$ \\
        \bottomrule
    \end{tabular}
    }
\end{table}

\begin{figure}
\centering
\subfloat[image quality]{
    \includegraphics[height=4cm]{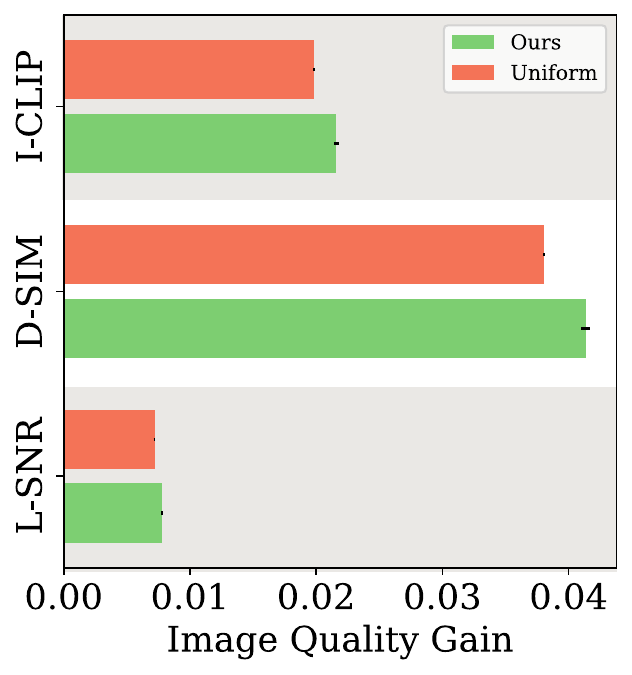}
    \label{fig:eval:gain}
  }
\subfloat[time consumption]{
    \includegraphics[height=4cm]{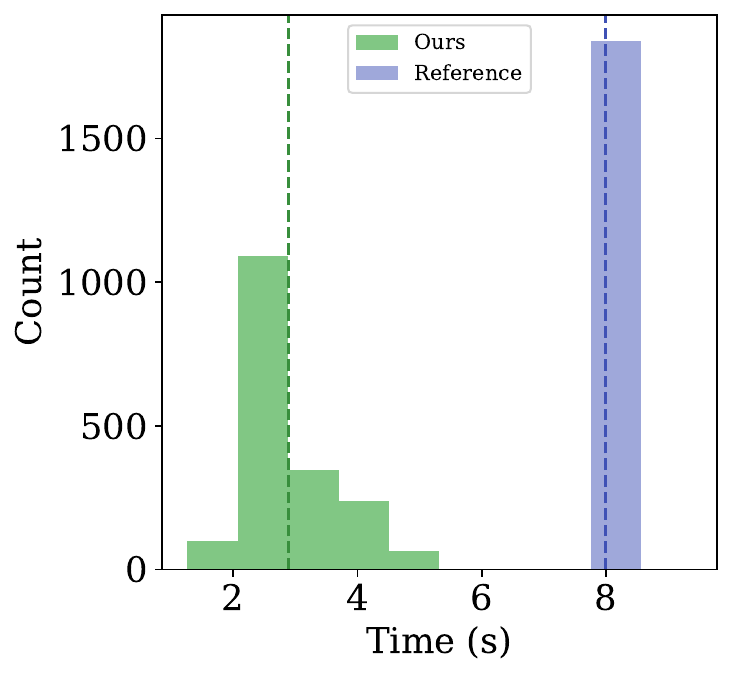}
    \label{fig:eval:time}
  }
\Caption{Quantitative results.}
{%
\subref{fig:eval:gain} We plot the average image quality gain, defined in \Cref{eq:gain}, of conditions \condOurs and \condUniform for the three perceptual metrics.
\subref{fig:eval:time} Average computation time (in seconds) with the corresponding number of forward denoising steps of conditions \condOurs and \condRef.
}
\label{fig:eval:hist}
\Description{}
\end{figure}

\paragraph{Overall quality gain}
The previous analysis suggests that \condOurs and \condUniform seem to exhibit comparable efficiency per step, and it may not be immediately clear that our technique is better than selecting a constant number of steps, as with the \condUniform condition.
Note that \condOurs and \condUniform share the same total number of denoising steps; we further measure their relative quality as the ratio between the two,
\begin{align}
    \frac{{\metric_{\step_\text{\condOurs}}(\prompt)}}{{\metric_{\step_\condUniform}(\prompt)}} - 1.
    \label{eqn:eval:ratio}
\end{align}
Here, $\step_\condOurs$ and $\step_\condUniform$ indicate the timestep numbers suggested by $\condOurs$ and $\condUniform$, respectively.
Intuitively, the positive or negative ratio values indicate how much \condOurs outperforms or underperforms than \condUniform. 
\Cref{fig:eval:gain} visualizes the results after computing average relative quality over all images in the test set.
This results in relative quality of $6.6\% \pm 0.8\%$ for $\metric$=\SNRLaplacian, $8.4\% \pm 0.9\%$ for $\metric$=\dreamsim, and $8.7\%\pm0.9\%$ for $\metric$=\clip.
These results show that, despite having the same total number of denoising timesteps (i.e., same total computation), \condOurs exhibits significant quality gain over \condUniform. This is also evidenced by our subjective study in the following \Cref{sec:eval:study}.

\nothing{
also measure the image quality of our method since only care about efficiency will lead to messy and noisy images. Therefore, we propose to compare the quality of images generated from our suggested number of steps to images generated from uniform numbers of steps whose average is similar to ours so that the average efficiency is similar. Then we do the t-test on the scores from three metrics improvement per step from the image generated with step 1 to the image generated with our suggested step and the image generated with uniform steps, and the Null hypothesis is that scores from the uniform steps are larger or equal to our suggested steps.

As we can see in the \Cref{tab:comparison}, the t-test fails with the confidence interval of 99 percent for both hypothesises and all types of scores. Therefore, we can conclude that the steps suggested by our method are efficient without compromising quality.
}

\paragraph{Time consumption}
To measure the tangible benefits between \condRef and \condOurs, we measure their time consumption as seconds per image (SPI) on an NVIDIA RTX 3090 GPU.
As shown in \Cref{fig:eval:time}, with the entire evaluation dataset, \condOurs achieved $2.89\pm 0.87$ SPI versus \condRef at $8.00 \pm 0.039$ SPI, which is $63.9\%$ of the time \nothing{(and therefore power) }of \condRef, indicating significantly reduced computational resource usage.



\subsection{Image Quality and Diversity}
\label{sec:eval:diversity}
We also compute the inception score (IS) to measure the generated image quality and diversity of the three conditions. The IS scores of \condOurs, \condUniform, and \condRef were 27.9, 27.9, and 28.7, respectively, which demonstrates that \methodName achieves computation savings without substantially compromising image quality and diversity.
Furthermore, we include visual results of our method \warning{in \Cref{fig:tileResults}, and }in the Supplementary Materials.



\subsection{Subjective Crowdsourcing Study}
\label{sec:eval:study}

\begin{figure}
\centering
\includegraphics[width=\linewidth]{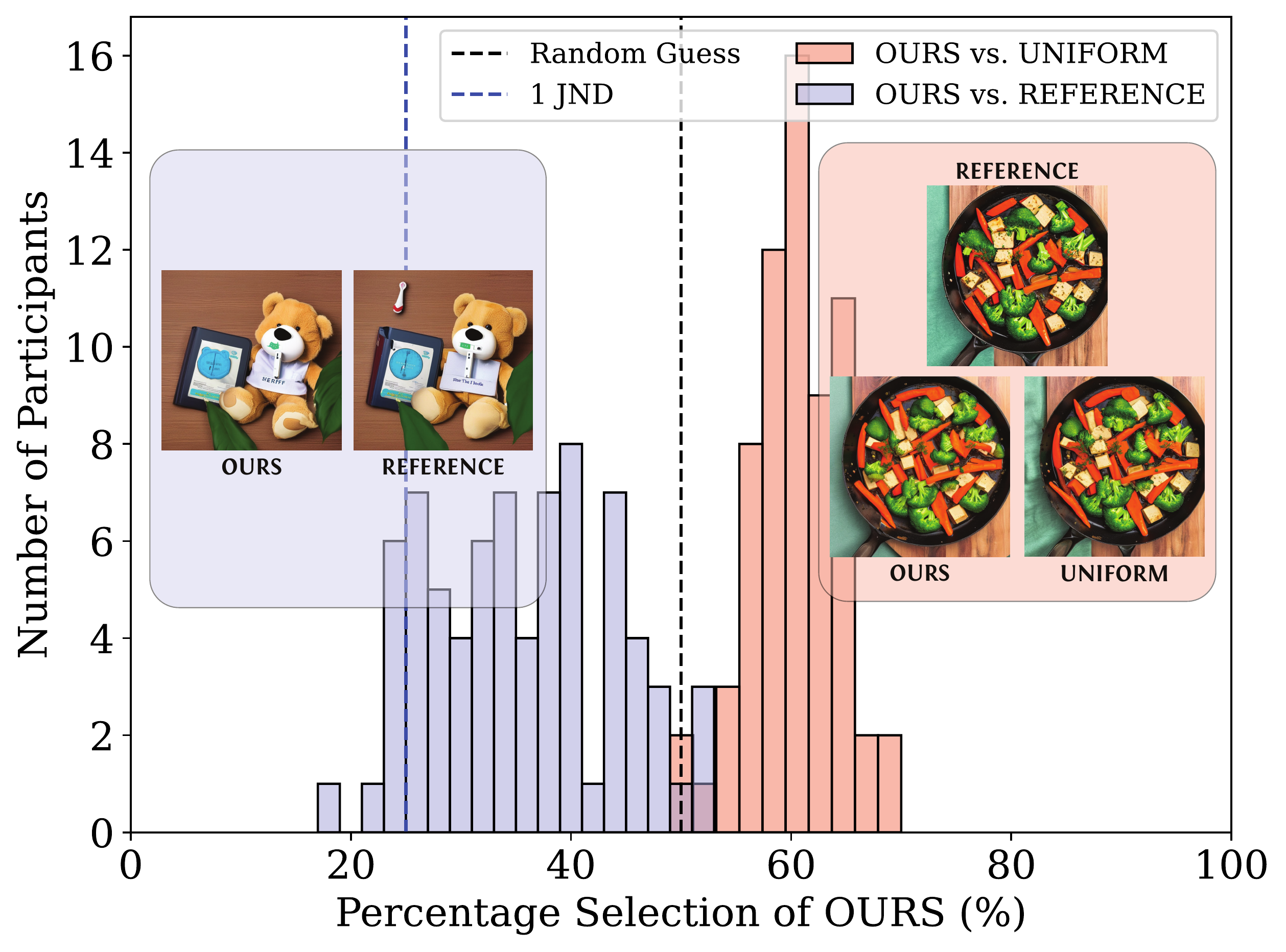}
\label{fig:study:results}
\Caption{Crowdsourced user study results.}
{
Here (top) we visualize user study results, with percentage selection on the $x-$axis and number of participants on the $y-$axis.
The green distribution are results for the \condOurs vs. \condUniform study, and the violet distribution is that of \condOurs vs. \condRef.
Random guess (50\%) is visualized as the black dashed line, and 1 JND threshold is the blue dashed line.
The insets show examples of presented stimuli to users for each study.
}
\label{fig:eval:user}
\Description{}
\end{figure}

As visualized in the insets of \Cref{fig:study:results}, we conducted a crowd-sourced user study with 72 participants through the platform \textit{Prolific} to determine whether images generated with our technique are of high quality.
\citet{mantiuk2012comparison} found that forced-choice studies require fewer trials, reduce variance in results, and provide an unambiguous task for participants.
As such, we ran a two-alternative forced-choice (2AFC) study to measure participants' preference for images generated by \condOurs and \condUniform.
During each trial, participants were presented with the reference and two test images, one generated by either \condOurs or \condUniform. Users were instructed to select the image with higher image quality, with respect to \condRef.
In total, participants completed 100 trials, with prompts randomly sampled from the dataset described in \cref{sec:method:data}.
An additional 2AFC study was conducted on 78 participants to compare the quality between \condRef and \condOurs.
Our study was approved by an Institutional Review Board (IRB).

\paragraph{Results and discussion}
The results are visualized in \Cref{fig:study:results} as a distribution of percentage selection of \condOurs.
Here, we define percentage selection as the proportion of trials in which users selected \condOurs over \condUniform or \condRef (depending on the study task).
The mean percentage of selection of \condOurs was $60.3\pm 3.9\%$.
In the study in which users compare \condOurs and \condRef, participants selected \condOurs $35.2\pm 8.4\%$ of the time, lower than 25\% selection which is commonly used to define the 1 Just-Noticeable-Difference (1 JND) threshold in psychophysical studies.
The 1 JND range is commonly used in prior literature attempting to determine perceptually unnoticeable threshold \cite{fu2023learning}. 
These results evidence that our \methodName-guided diffusion model achieves significant computation savings without sacrificing subjectively perceived quality.
It may be interesting, in future work, to determine a mapping from plateau computation (e.g., defined in \cref{sec:method:score2step}) to percentage preference.
We expand on this discussion in the next section.


\section{Limitations and Future Work \complete}
\label{sec:future-work}

\paragraph{Sparsely sampled diffusion timesteps}
While preparing the dataset in \Cref{sec:method:data}, we sample the timesteps $\stepSet$ in a non-uniform fashion considering the perceptual power law and then linearly interpolate the perceptual metrics among them during model training (\Cref{sec:method:text2score}). 
This is due to the storage and computation efficiency. As a future follow-up, evaluating perceptual quality with dense and uniform sampling may provide more fine-grained guidance.


\paragraph{Variable savings.}
In this work, we develop an algorithm to compute the plateau point of three metrics (see \cref{sec:method:score2step}), which is used to determine a dataset for driving our technique. 
Ultimately, optimal timesteps determined by \condOurs depends on this dataset.
It may be useful to determine how this dataset could be modified, for example in order to select optimal timesteps which require a pre-defined amount of compute.

\paragraph{Extension to videos.} \methodName focuses on optimizing the quality-computation efficiency for image generation. More recently, significant advancements have been made to high-fidelity video generation \cite{guo2023animatediff,ho2022video,croitoru2023diffusion}. Accessing the perceptual quality in the spatio-temporal domain would require a substantial amount of data and perceptual measurements \cite{chikkerur2011objective,mantiuk2021fovvideovdp}, and therefore beyond the scope of this research as the first attempt to guide diffusion models by human visual perception. 
As an exciting future direction, we envision solutions such as latent feature extraction \cite{sun2023spatio} may shed light on reducing the required data volume for robust guidance.


\section{Conclusion\complete}
In this paper, we present \methodName, a predictive perceptual guidance for diffusion models. Given a text prompt, it suggests the most efficient timesteps \textit{before} the computationally intensive neural-network-based denoising diffusion starts. That is, it guides the diffusion toward optimized ``perceptual gain per step''.
Our objective analysis and user studies show that this is achieved without compromising perceptual quality and content diversity. 
As ``\textit{making an image with generative AI uses as much energy as charging your phone}'' \cite{genaiIphone}, we hope this research will pave the way for future human-centered approaches to address the surging questions of how to optimize the computational and emission costs of generative models.

{
    \small
    \bibliographystyle{ieeenat_fullname}
    \bibliography{main}

\begin{thebibliography}{60}
\providecommand{\natexlab}[1]{#1}
\providecommand{\url}[1]{\texttt{#1}}
\expandafter\ifx\csname urlstyle\endcsname\relax
  \providecommand{\doi}[1]{doi: #1}\else
  \providecommand{\doi}{doi: \begingroup \urlstyle{rm}\Url}\fi

\bibitem[Barraco et~al.(2022)Barraco, Cornia, Cascianelli, Baraldi, and Cucchiara]{barraco2022unreasonable}
Manuele Barraco, Marcella Cornia, Silvia Cascianelli, Lorenzo Baraldi, and Rita Cucchiara.
\newblock The unreasonable effectiveness of clip features for image captioning: an experimental analysis.
\newblock In \emph{proceedings of the IEEE/CVF conference on computer vision and pattern recognition}, pages 4662--4670, 2022.

\bibitem[Chen et~al.(2023)Chen, Sarokin, Lee, Tang, Chang, Kulik, and Grundmann]{chen2023speed}
Yu-Hui Chen, Raman Sarokin, Juhyun Lee, Jiuqiang Tang, Chuo-Ling Chang, Andrei Kulik, and Matthias Grundmann.
\newblock Speed is all you need: On-device acceleration of large diffusion models via gpu-aware optimizations.
\newblock In \emph{Proceedings of the IEEE/CVF Conference on Computer Vision and Pattern Recognition}, pages 4650--4654, 2023.

\bibitem[Cheragui et~al.(2023)Cheragui, Dahou, and Abdedaiem]{cheragui2023exploring}
Mohamed~Amine Cheragui, Abdelhalim~Hafedh Dahou, and Amin Abdedaiem.
\newblock Exploring bert models for part-of-speech tagging in the algerian dialect: A comprehensive study.
\newblock In \emph{Proceedings of the 6th International Conference on Natural Language and Speech Processing (ICNLSP 2023)}, pages 140--150, 2023.

\bibitem[Chikkerur et~al.(2011)Chikkerur, Sundaram, Reisslein, and Karam]{chikkerur2011objective}
Shyamprasad Chikkerur, Vijay Sundaram, Martin Reisslein, and Lina~J Karam.
\newblock Objective video quality assessment methods: A classification, review, and performance comparison.
\newblock \emph{IEEE transactions on broadcasting}, 57\penalty0 (2):\penalty0 165--182, 2011.

\bibitem[Croitoru et~al.(2023)Croitoru, Hondru, Ionescu, and Shah]{croitoru2023diffusion}
Florinel-Alin Croitoru, Vlad Hondru, Radu~Tudor Ionescu, and Mubarak Shah.
\newblock Diffusion models in vision: A survey.
\newblock \emph{IEEE Transactions on Pattern Analysis and Machine Intelligence}, 2023.

\bibitem[Czolbe et~al.(2020)Czolbe, Krause, Cox, and Igel]{czolbe2020loss}
Steffen Czolbe, Oswin Krause, Ingemar Cox, and Christian Igel.
\newblock A loss function for generative neural networks based on watson’s perceptual model.
\newblock \emph{Advances in Neural Information Processing Systems}, 33:\penalty0 2051--2061, 2020.

\bibitem[Daly(1992)]{daly1992visible}
Scott~J Daly.
\newblock Visible differences predictor: an algorithm for the assessment of image fidelity.
\newblock In \emph{Human Vision, Visual Processing, and Digital Display III}, pages 2--15. SPIE, 1992.

\bibitem[Engbert and Mergenthaler(2006)]{engbert2006microsaccades}
Ralf Engbert and Konstantin Mergenthaler.
\newblock Microsaccades are triggered by low retinal image slip.
\newblock \emph{Proceedings of the National Academy of Sciences}, 103\penalty0 (18):\penalty0 7192--7197, 2006.

\bibitem[Fu et~al.(2024)Fu, Tamir, Sundaram, Chai, Zhang, Dekel, and Isola]{fu2023learning}
Stephanie Fu, Netanel Tamir, Shobhita Sundaram, Lucy Chai, Richard Zhang, Tali Dekel, and Phillip Isola.
\newblock Dreamsim: Learning new dimensions of human visual similarity using synthetic data.
\newblock \emph{Advances in Neural Information Processing Systems}, 36, 2024.

\bibitem[Graves and Schmidhuber(2005)]{graves2005framewise}
Alex Graves and J{\"u}rgen Schmidhuber.
\newblock Framewise phoneme classification with bidirectional lstm and other neural network architectures.
\newblock \emph{Neural networks}, 18\penalty0 (5-6):\penalty0 602--610, 2005.

\bibitem[Graves et~al.(2013)Graves, Mohamed, and Hinton]{graves2013speech}
Alex Graves, Abdel-rahman Mohamed, and Geoffrey Hinton.
\newblock Speech recognition with deep recurrent neural networks.
\newblock In \emph{2013 IEEE international conference on acoustics, speech and signal processing}, pages 6645--6649. Ieee, 2013.

\bibitem[Guo et~al.(2023)Guo, Yang, Rao, Wang, Qiao, Lin, and Dai]{guo2023animatediff}
Yuwei Guo, Ceyuan Yang, Anyi Rao, Yaohui Wang, Yu Qiao, Dahua Lin, and Bo Dai.
\newblock Animatediff: Animate your personalized text-to-image diffusion models without specific tuning.
\newblock \emph{arXiv preprint arXiv:2307.04725}, 2023.

\bibitem[Hinton et~al.(2012)Hinton, Srivastava, Krizhevsky, Sutskever, and Salakhutdinov]{hinton2012improving}
Geoffrey~E Hinton, Nitish Srivastava, Alex Krizhevsky, Ilya Sutskever, and Ruslan~R Salakhutdinov.
\newblock Improving neural networks by preventing co-adaptation of feature detectors.
\newblock \emph{arXiv preprint arXiv:1207.0580}, 2012.

\bibitem[Ho et~al.(2020)Ho, Jain, and Abbeel]{ho2020denoising}
Jonathan Ho, Ajay Jain, and Pieter Abbeel.
\newblock Denoising diffusion probabilistic models.
\newblock \emph{Advances in neural information processing systems}, 33:\penalty0 6840--6851, 2020.

\bibitem[Ho et~al.(2022)Ho, Salimans, Gritsenko, Chan, Norouzi, and Fleet]{ho2022video}
Jonathan Ho, Tim Salimans, Alexey Gritsenko, William Chan, Mohammad Norouzi, and David~J Fleet.
\newblock Video diffusion models.
\newblock \emph{Advances in Neural Information Processing Systems}, 35:\penalty0 8633--8646, 2022.

\bibitem[Jindal et~al.(2021)Jindal, Wolski, Myszkowski, and Mantiuk]{jindal2021perceptual}
Akshay Jindal, Krzysztof Wolski, Karol Myszkowski, and Rafa{\l}~K Mantiuk.
\newblock Perceptual model for adaptive local shading and refresh rate.
\newblock \emph{ACM Transactions on Graphics (TOG)}, 40\penalty0 (6):\penalty0 1--18, 2021.

\bibitem[Johnson et~al.(2016)Johnson, Alahi, and Fei-Fei]{johnson2016perceptual}
Justin Johnson, Alexandre Alahi, and Li Fei-Fei.
\newblock Perceptual losses for real-time style transfer and super-resolution.
\newblock In \emph{Computer Vision--ECCV 2016: 14th European Conference, Amsterdam, The Netherlands, October 11-14, 2016, Proceedings, Part II 14}, pages 694--711. Springer, 2016.

\bibitem[Kaack et~al.(2022)Kaack, Donti, Strubell, Kamiya, Creutzig, and Rolnick]{kaack2022aligning}
Lynn~H Kaack, Priya~L Donti, Emma Strubell, George Kamiya, Felix Creutzig, and David Rolnick.
\newblock Aligning artificial intelligence with climate change mitigation.
\newblock \emph{Nature Climate Change}, 12\penalty0 (6):\penalty0 518--527, 2022.

\bibitem[Karras et~al.(2022)Karras, Aittala, Aila, and Laine]{karras2022elucidating}
Tero Karras, Miika Aittala, Timo Aila, and Samuli Laine.
\newblock Elucidating the design space of diffusion-based generative models.
\newblock \emph{Advances in Neural Information Processing Systems}, 35:\penalty0 26565--26577, 2022.

\bibitem[{Kate Crawford}(2024)]{genaiEnergyNature}
{Kate Crawford}.
\newblock Generative ai’s environmental costs are soaring — and mostly secret.
\newblock \emph{Nature World View}, 2024.

\bibitem[Kellman and McVeigh(2005)]{kellman2005image}
Peter Kellman and Elliot~R McVeigh.
\newblock Image reconstruction in snr units: a general method for snr measurement.
\newblock \emph{Magnetic resonance in medicine}, 54\penalty0 (6):\penalty0 1439--1447, 2005.

\bibitem[Kenton and Toutanova(2019)]{kenton2019bert}
Jacob Devlin Ming-Wei~Chang Kenton and Lee~Kristina Toutanova.
\newblock Bert: Pre-training of deep bidirectional transformers for language understanding.
\newblock In \emph{Proceedings of NAACL-HLT}, pages 4171--4186, 2019.

\bibitem[Li et~al.(2023)Li, Li, Zheng, Wu, Xiao, Wang, Zheng, Pan, Chao, and Ji]{li2023autodiffusion}
Lijiang Li, Huixia Li, Xiawu Zheng, Jie Wu, Xuefeng Xiao, Rui Wang, Min Zheng, Xin Pan, Fei Chao, and Rongrong Ji.
\newblock Autodiffusion: Training-free optimization of time steps and architectures for automated diffusion model acceleration.
\newblock In \emph{Proceedings of the IEEE/CVF International Conference on Computer Vision}, pages 7105--7114, 2023.

\bibitem[Li et~al.(2024{\natexlab{a}})Li, Cai, Cao, Zhang, Cai, Bai, Jia, Li, and Han]{li2024distrifusion}
Muyang Li, Tianle Cai, Jiaxin Cao, Qinsheng Zhang, Han Cai, Junjie Bai, Yangqing Jia, Kai Li, and Song Han.
\newblock Distrifusion: Distributed parallel inference for high-resolution diffusion models.
\newblock In \emph{Proceedings of the IEEE/CVF Conference on Computer Vision and Pattern Recognition}, pages 7183--7193, 2024{\natexlab{a}}.

\bibitem[Li et~al.(2024{\natexlab{b}})Li, Wang, Jin, Hu, Chemerys, Fu, Wang, Tulyakov, and Ren]{li2024snapfusion}
Yanyu Li, Huan Wang, Qing Jin, Ju Hu, Pavlo Chemerys, Yun Fu, Yanzhi Wang, Sergey Tulyakov, and Jian Ren.
\newblock Snapfusion: Text-to-image diffusion model on mobile devices within two seconds.
\newblock \emph{Advances in Neural Information Processing Systems}, 36, 2024{\natexlab{b}}.

\bibitem[Lin et~al.(2014)Lin, Maire, Belongie, Hays, Perona, Ramanan, Doll{\'a}r, and Zitnick]{lin2014microsoft}
Tsung-Yi Lin, Michael Maire, Serge Belongie, James Hays, Pietro Perona, Deva Ramanan, Piotr Doll{\'a}r, and C~Lawrence Zitnick.
\newblock Microsoft coco: Common objects in context.
\newblock In \emph{Computer Vision--ECCV 2014: 13th European Conference, Zurich, Switzerland, September 6-12, 2014, Proceedings, Part V 13}, pages 740--755. Springer, 2014.

\bibitem[Liu et~al.(2022)Liu, Ren, Lin, and Zhao]{liu2022pseudo}
Luping Liu, Yi Ren, Zhijie Lin, and Zhou Zhao.
\newblock Pseudo numerical methods for diffusion models on manifolds.
\newblock \emph{arXiv preprint arXiv:2202.09778}, 2022.

\bibitem[Liu et~al.(2023)Liu, Zhang, Ma, Peng, et~al.]{liu2023instaflow}
Xingchao Liu, Xiwen Zhang, Jianzhu Ma, Jian Peng, et~al.
\newblock Instaflow: One step is enough for high-quality diffusion-based text-to-image generation.
\newblock In \emph{The Twelfth International Conference on Learning Representations}, 2023.

\bibitem[Lu et~al.(2022)Lu, Zhou, Bao, Chen, Li, and Zhu]{lu2022dpm}
Cheng Lu, Yuhao Zhou, Fan Bao, Jianfei Chen, Chongxuan Li, and Jun Zhu.
\newblock Dpm-solver: A fast ode solver for diffusion probabilistic model sampling in around 10 steps.
\newblock \emph{Advances in Neural Information Processing Systems}, 35:\penalty0 5775--5787, 2022.

\bibitem[Luo et~al.(2023)Luo, Tan, Huang, Li, and Zhao]{luo2023latent}
Simian Luo, Yiqin Tan, Longbo Huang, Jian Li, and Hang Zhao.
\newblock Latent consistency models: Synthesizing high-resolution images with few-step inference.
\newblock \emph{arXiv preprint arXiv:2310.04378}, 2023.

\bibitem[Ma et~al.(2024)Ma, Fang, and Wang]{ma2023deepcache}
Xinyin Ma, Gongfan Fang, and Xinchao Wang.
\newblock Deepcache: Accelerating diffusion models for free.
\newblock In \emph{The IEEE/CVF Conference on Computer Vision and Pattern Recognition}, 2024.

\bibitem[Mahadevaswamy and Swathi(2023)]{mahadevaswamy2023sentiment}
UB Mahadevaswamy and P Swathi.
\newblock Sentiment analysis using bidirectional lstm network.
\newblock \emph{Procedia Computer Science}, 218:\penalty0 45--56, 2023.

\bibitem[Mantiuk et~al.(2011)Mantiuk, Kim, Rempel, and Heidrich]{mantiuk2011hdr}
Rafa{\l} Mantiuk, Kil~Joong Kim, Allan~G Rempel, and Wolfgang Heidrich.
\newblock Hdr-vdp-2: A calibrated visual metric for visibility and quality predictions in all luminance conditions.
\newblock \emph{ACM Transactions on graphics (TOG)}, 30\penalty0 (4):\penalty0 1--14, 2011.

\bibitem[Mantiuk et~al.(2012)Mantiuk, Tomaszewska, and Mantiuk]{mantiuk2012comparison}
Rafa{\l}~K Mantiuk, Anna Tomaszewska, and Rados{\l}aw Mantiuk.
\newblock Comparison of four subjective methods for image quality assessment.
\newblock In \emph{Computer graphics forum}, pages 2478--2491. Wiley Online Library, 2012.

\bibitem[Mantiuk et~al.(2021)Mantiuk, Denes, Chapiro, Kaplanyan, Rufo, Bachy, Lian, and Patney]{mantiuk2021fovvideovdp}
Rafa{\l}~K Mantiuk, Gyorgy Denes, Alexandre Chapiro, Anton Kaplanyan, Gizem Rufo, Romain Bachy, Trisha Lian, and Anjul Patney.
\newblock Fovvideovdp: A visible difference predictor for wide field-of-view video.
\newblock \emph{ACM Transactions on Graphics (TOG)}, 40\penalty0 (4):\penalty0 1--19, 2021.

\bibitem[{Melissa Heikkilä}(2023)]{genaiIphone}
{Melissa Heikkilä}.
\newblock Making an image with generative ai uses as much energy as charging your phone.
\newblock \emph{MIT Technology Review}, 2023.

\bibitem[Meng et~al.(2023)Meng, Rombach, Gao, Kingma, Ermon, Ho, and Salimans]{meng2023distillation}
Chenlin Meng, Robin Rombach, Ruiqi Gao, Diederik Kingma, Stefano Ermon, Jonathan Ho, and Tim Salimans.
\newblock On distillation of guided diffusion models.
\newblock In \emph{Proceedings of the IEEE/CVF Conference on Computer Vision and Pattern Recognition}, pages 14297--14306, 2023.

\bibitem[Mohammad~Khalid et~al.(2022)Mohammad~Khalid, Xie, Belilovsky, and Popa]{mohammad2022clip}
Nasir Mohammad~Khalid, Tianhao Xie, Eugene Belilovsky, and Tiberiu Popa.
\newblock Clip-mesh: Generating textured meshes from text using pretrained image-text models.
\newblock In \emph{SIGGRAPH Asia 2022 conference papers}, pages 1--8, 2022.

\bibitem[Nguyen et~al.(2024)Nguyen, Vu, Thanh, Than, and Tran]{nguyen2024inference}
Viet Nguyen, Giang Vu, Tung~Nguyen Thanh, Khoat Than, and Toan Tran.
\newblock On inference stability for diffusion models.
\newblock In \emph{Proceedings of the AAAI Conference on Artificial Intelligence}, pages 14449--14456, 2024.

\bibitem[Nichol and Dhariwal(2021)]{nichol2021improved}
Alexander~Quinn Nichol and Prafulla Dhariwal.
\newblock Improved denoising diffusion probabilistic models.
\newblock In \emph{International conference on machine learning}, pages 8162--8171. PMLR, 2021.

\bibitem[Oliveira et~al.(2020)Oliveira, Moussallem, Garcia, and Fileto]{oliveira2020optic}
Italo~Lopes Oliveira, Diego Moussallem, Lu{\'\i}s Paulo~Faina Garcia, and Renato Fileto.
\newblock Optic: A deep neural network approach for entity linking using word and knowledge embeddings.
\newblock In \emph{ICEIS (1)}, pages 315--326, 2020.

\bibitem[Paris et~al.(2011)Paris, Hasinoff, and Kautz]{paris2011local}
Sylvain Paris, Samuel~W Hasinoff, and Jan Kautz.
\newblock Local laplacian filters: Edge-aware image processing with a laplacian pyramid.
\newblock \emph{ACM Trans. Graph.}, 30\penalty0 (4):\penalty0 68, 2011.

\bibitem[Patney et~al.(2016)Patney, Salvi, Kim, Kaplanyan, Wyman, Benty, Luebke, and Lefohn]{patney2016towards}
Anjul Patney, Marco Salvi, Joohwan Kim, Anton Kaplanyan, Chris Wyman, Nir Benty, David Luebke, and Aaron Lefohn.
\newblock Towards foveated rendering for gaze-tracked virtual reality.
\newblock \emph{ACM Transactions on Graphics (TOG)}, 35\penalty0 (6):\penalty0 1--12, 2016.

\bibitem[Preechakul et~al.(2022)Preechakul, Chatthee, Wizadwongsa, and Suwajanakorn]{preechakul2022diffusion}
Konpat Preechakul, Nattanat Chatthee, Suttisak Wizadwongsa, and Supasorn Suwajanakorn.
\newblock Diffusion autoencoders: Toward a meaningful and decodable representation.
\newblock In \emph{Proceedings of the IEEE/CVF Conference on Computer Vision and Pattern Recognition}, pages 10619--10629, 2022.

\bibitem[Radford et~al.(2021)Radford, Kim, Hallacy, Ramesh, Goh, Agarwal, Sastry, Askell, Mishkin, Clark, et~al.]{radford2021learning}
Alec Radford, Jong~Wook Kim, Chris Hallacy, Aditya Ramesh, Gabriel Goh, Sandhini Agarwal, Girish Sastry, Amanda Askell, Pamela Mishkin, Jack Clark, et~al.
\newblock Learning transferable visual models from natural language supervision.
\newblock In \emph{International conference on machine learning}, pages 8748--8763. PMLR, 2021.

\bibitem[Ramesh et~al.(2021)Ramesh, Pavlov, Goh, Gray, Voss, Radford, Chen, and Sutskever]{ramesh2021zero}
Aditya Ramesh, Mikhail Pavlov, Gabriel Goh, Scott Gray, Chelsea Voss, Alec Radford, Mark Chen, and Ilya Sutskever.
\newblock Zero-shot text-to-image generation.
\newblock In \emph{International conference on machine learning}, pages 8821--8831. Pmlr, 2021.

\bibitem[Rombach et~al.(2022)Rombach, Blattmann, Lorenz, Esser, and Ommer]{rombach2022high}
Robin Rombach, Andreas Blattmann, Dominik Lorenz, Patrick Esser, and Bj{\"o}rn Ommer.
\newblock High-resolution image synthesis with latent diffusion models.
\newblock In \emph{Proceedings of the IEEE/CVF conference on computer vision and pattern recognition}, pages 10684--10695, 2022.

\bibitem[Salimans and Ho(2022)]{salimans2022progressive}
Tim Salimans and Jonathan Ho.
\newblock Progressive distillation for fast sampling of diffusion models.
\newblock \emph{arXiv preprint arXiv:2202.00512}, 2022.

\bibitem[{Sarah Wells}(2023)]{genaiEnergy}
{Sarah Wells}.
\newblock Generative ai's energy problem today is foundational.
\newblock \emph{IEEE Spectrum}, 2023.

\bibitem[Sauer et~al.(2024)Sauer, Boesel, Dockhorn, Blattmann, Esser, and Rombach]{sauer2024fast}
Axel Sauer, Frederic Boesel, Tim Dockhorn, Andreas Blattmann, Patrick Esser, and Robin Rombach.
\newblock Fast high-resolution image synthesis with latent adversarial diffusion distillation.
\newblock \emph{arXiv preprint arXiv:2403.12015}, 2024.

\bibitem[Siami-Namini et~al.(2019)Siami-Namini, Tavakoli, and Namin]{siami2019performance}
Sima Siami-Namini, Neda Tavakoli, and Akbar~Siami Namin.
\newblock The performance of lstm and bilstm in forecasting time series.
\newblock In \emph{2019 IEEE International conference on big data (Big Data)}, pages 3285--3292. IEEE, 2019.

\bibitem[Song et~al.(2020)Song, Meng, and Ermon]{song2020denoising}
Jiaming Song, Chenlin Meng, and Stefano Ermon.
\newblock Denoising diffusion implicit models.
\newblock \emph{arXiv preprint arXiv:2010.02502}, 2020.

\bibitem[Song et~al.(2023)Song, Dhariwal, Chen, and Sutskever]{song2023consistency}
Yang Song, Prafulla Dhariwal, Mark Chen, and Ilya Sutskever.
\newblock Consistency models.
\newblock \emph{arXiv preprint arXiv:2303.01469}, 2023.

\bibitem[Souza and Filho(2022)]{souza2022bert}
Frederico~Dias Souza and Jo{\~a}o Baptista de Oliveira e~Souza Filho.
\newblock Bert for sentiment analysis: pre-trained and fine-tuned alternatives.
\newblock In \emph{International Conference on Computational Processing of the Portuguese Language}, pages 209--218. Springer, 2022.

\bibitem[Sun et~al.(2023)Sun, Wang, Zhang, Deng, Zafeiriou, and Hua]{sun2023spatio}
Guanxiong Sun, Chi Wang, Zhaoyu Zhang, Jiankang Deng, Stefanos Zafeiriou, and Yang Hua.
\newblock Spatio-temporal prompting network for robust video feature extraction.
\newblock In \emph{Proceedings of the IEEE/CVF International Conference on Computer Vision}, pages 13587--13597, 2023.

\bibitem[Wang et~al.(2004)Wang, Bovik, Sheikh, and Simoncelli]{wang2004image}
Zhou Wang, Alan~C Bovik, Hamid~R Sheikh, and Eero~P Simoncelli.
\newblock Image quality assessment: from error visibility to structural similarity.
\newblock \emph{IEEE transactions on image processing}, 13\penalty0 (4):\penalty0 600--612, 2004.

\bibitem[Wang et~al.(2022)Wang, Liu, He, Wu, and Yi]{wang2022clip}
Zihao Wang, Wei Liu, Qian He, Xinglong Wu, and Zili Yi.
\newblock Clip-gen: Language-free training of a text-to-image generator with clip.
\newblock \emph{arXiv preprint arXiv:2203.00386}, 2022.

\bibitem[Yang et~al.(2023)Yang, Wang, Qian, Zhu, and Wu]{yang2023diffusion}
Yongqi Yang, Ruoyu Wang, Zhihao Qian, Ye Zhu, and Yu Wu.
\newblock Diffusion in diffusion: Cyclic one-way diffusion for text-vision-conditioned generation.
\newblock \emph{arXiv preprint arXiv:2306.08247}, 2023.

\bibitem[Yuan et~al.(2019)Yuan, Deng, Tang, Tang, and Chen]{yuan2019signal}
Tongtong Yuan, Weihong Deng, Jian Tang, Yinan Tang, and Binghui Chen.
\newblock Signal-to-noise ratio: A robust distance metric for deep metric learning.
\newblock In \emph{Proceedings of the IEEE/CVF conference on computer vision and pattern recognition}, pages 4815--4824, 2019.

\bibitem[Zhang et~al.(2018)Zhang, Isola, Efros, Shechtman, and Wang]{zhang2018unreasonable}
Richard Zhang, Phillip Isola, Alexei~A Efros, Eli Shechtman, and Oliver Wang.
\newblock The unreasonable effectiveness of deep features as a perceptual metric.
\newblock In \emph{Proceedings of the IEEE conference on computer vision and pattern recognition}, pages 586--595, 2018.

\end{thebibliography}
}





\onecolumn

\setcounter{page}{1}
{
\centering
\Large
\textbf{\thetitle}\\
\vspace{0.5em}Supplementary Material \\
\vspace{1.0em}
}
\section*{Additional Model Training Details} 
We used one A100 to train models, and  ~30 minutes for \SNRLaplacian and \dreamsim models, and ~ 60 minutes for the \clip model. In inference, our model takes ~0.004 seconds to predict quality scores for a given prompt. \qinchan{perhaps we do not have to mention additional details since we have most of them in the method section.}

\section*{User Study Protocol}
\begin{figure*}[h!]
\centering
\includegraphics[width=\linewidth]{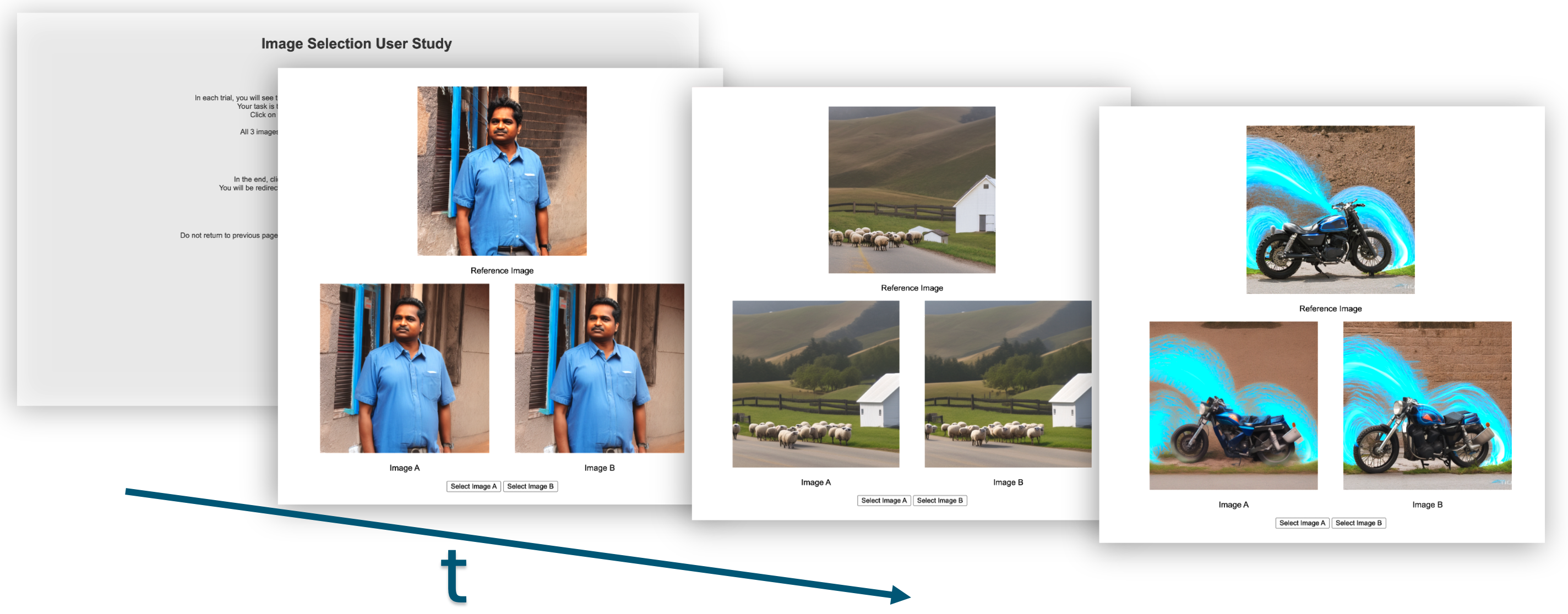}
\Caption{User study protocol OURS vs. UNIFORM.}
{
In each user study trial, the participant will see three images: a reference image in the middle, and 2 test images (Image A and Image B). The task is to select the image that is more similar to the reference image. The participant needs to click on the button below or press the keyboard to choose A/B.
}
\label{fig:supp:protocol1}
\Description{}
\end{figure*}

\begin{figure*}[h!]
\centering
\includegraphics[width=\linewidth]{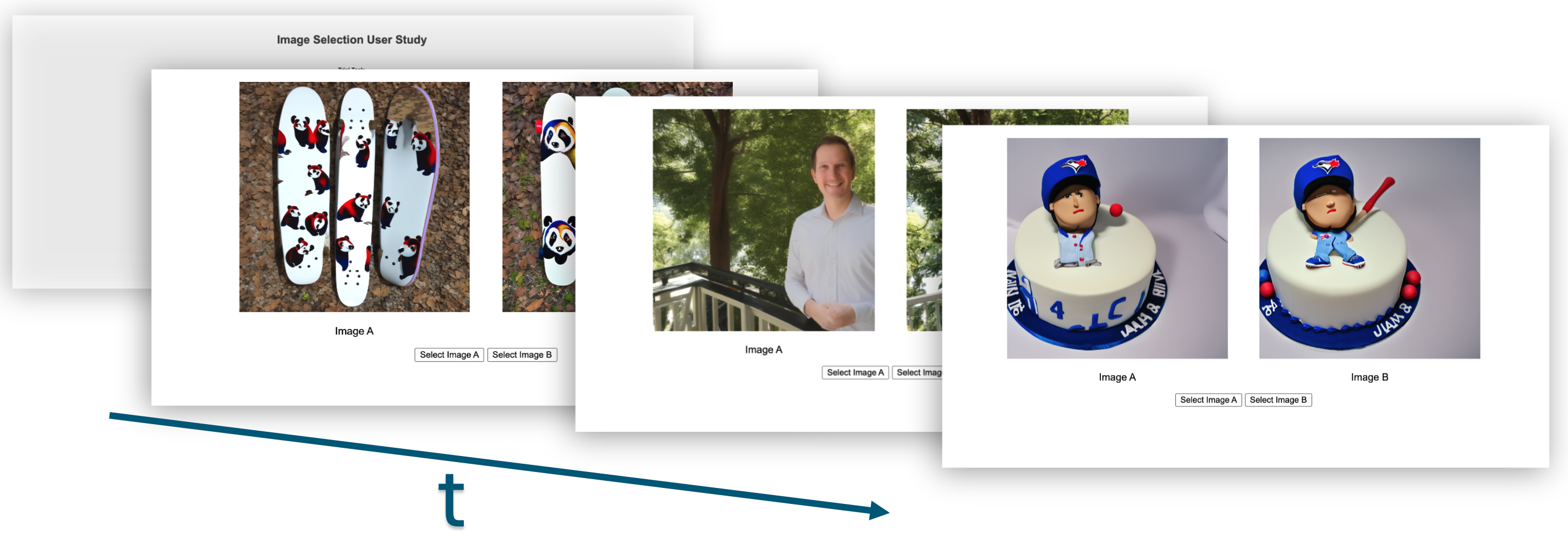}
\Caption{User study protocol OURS vs. REFERENCE.}
{
In each trial, the participant will see two images: Image A and Image B.
One of these two images is the reference image, which is of higher quality.
The participants need to choose the one they think is the reference image.
They will click on the button below or press the keyboard to choose A/B.
}
\label{fig:supp:protocol2}
\Description{}
\end{figure*}

Figure \ref{fig:supp:protocol1} and \ref{fig:supp:protocol2} visualize our crowdsourced user study protocol as a time sequence and example stimuli. 

\clearpage
\section*{Visualization of Evaluation Errors}
\begin{figure}[h!]
\centering
\subfloat[\SNRLaplacian error]{%
    \includegraphics[width=0.9\linewidth]{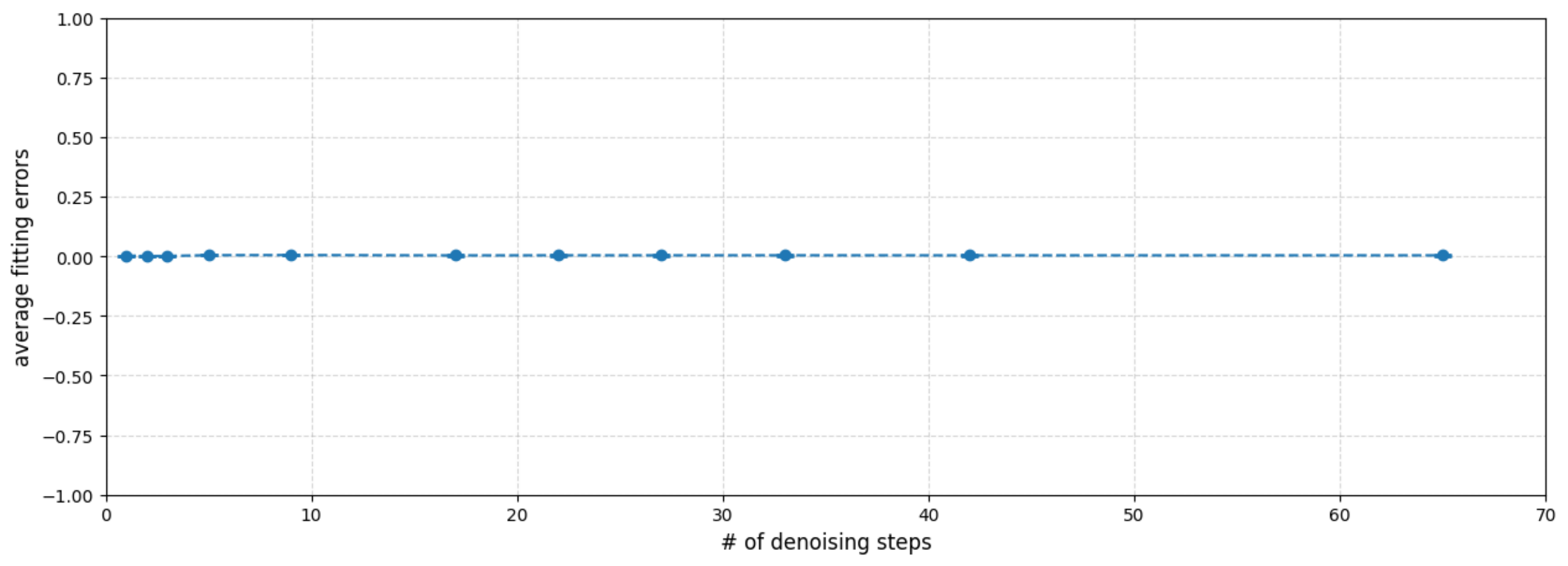}
    \label{fig:fittingerror:SNR}%
  }
\newline
\subfloat[\dreamsim error]{%
    \includegraphics[width=0.9\linewidth]{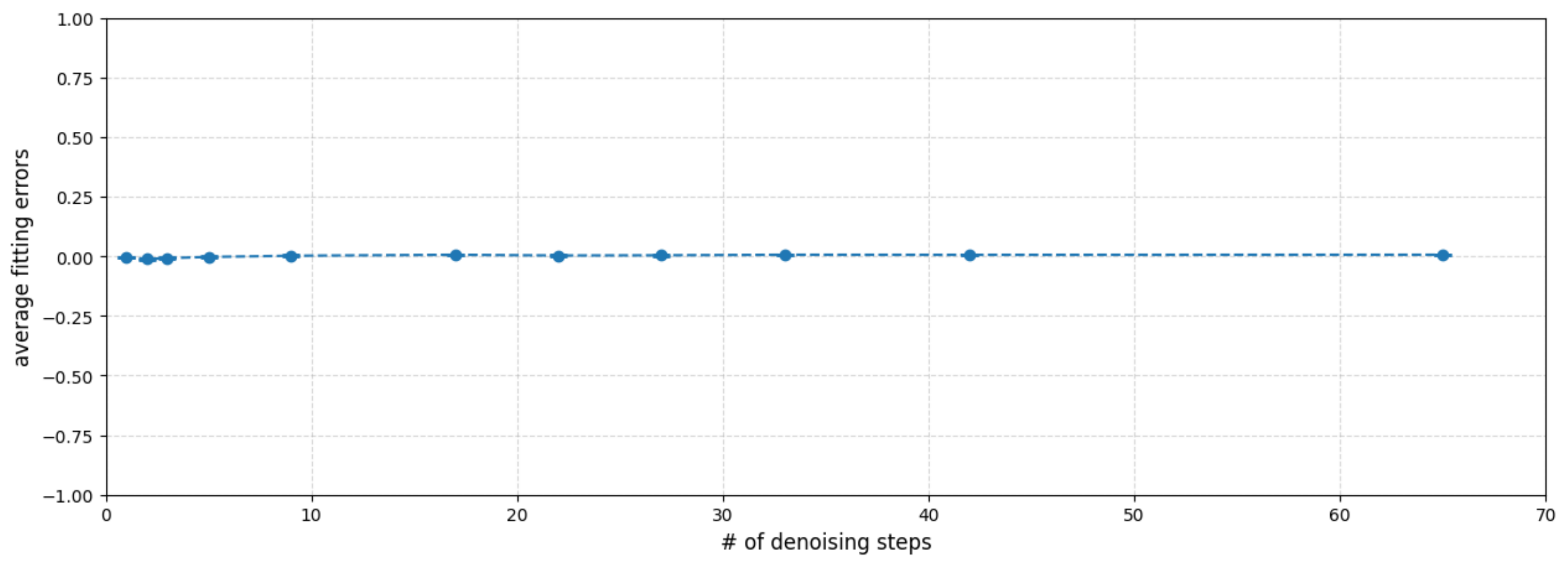}
    \label{fig:fittingerror:DreamSim}%
  }
\newline
\subfloat[\clip error]{%
    \includegraphics[width=0.9\linewidth]{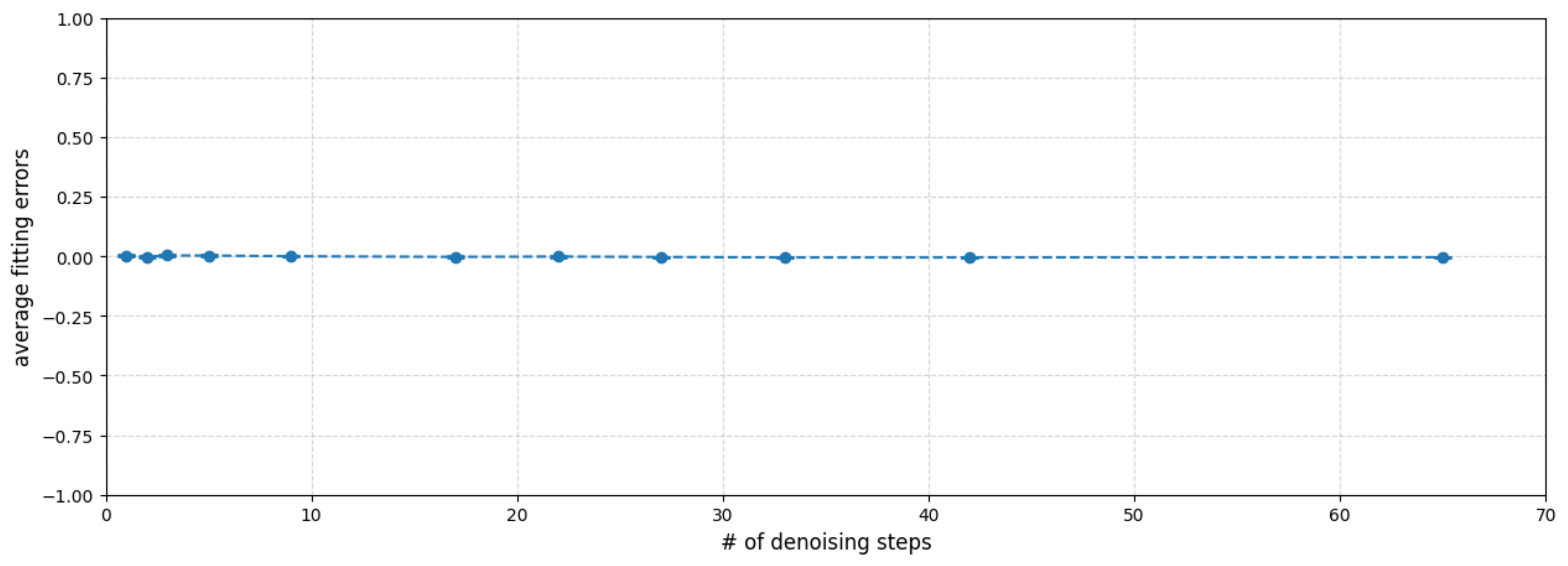}
    \label{fig:fittingerror:DreamSim}%
  }
\Caption{Visualizing fitting errors.}%
{$x$- and $y$-axis denotes the model-suggested number of denoising steps, and the corresponding fitting errors, respectively. The error bars indicate standard error. The results evidence the high accuracy and consistency of our LSTM prediction.}
\Description{}
\label{fig:fittingerror}
\end{figure}

Figure \ref{fig:fittingerror} visualizes the fitting error of our LSTM model across all three perceptual metrics.
For all number of denoising steps, error is low. 
This shows that our model is consistent across different number of timesteps.

\clearpage
\section*{Additional Qualitative Results}
\begin{figure*}[h!]
  \centering
  \includegraphics[width=0.98\linewidth]{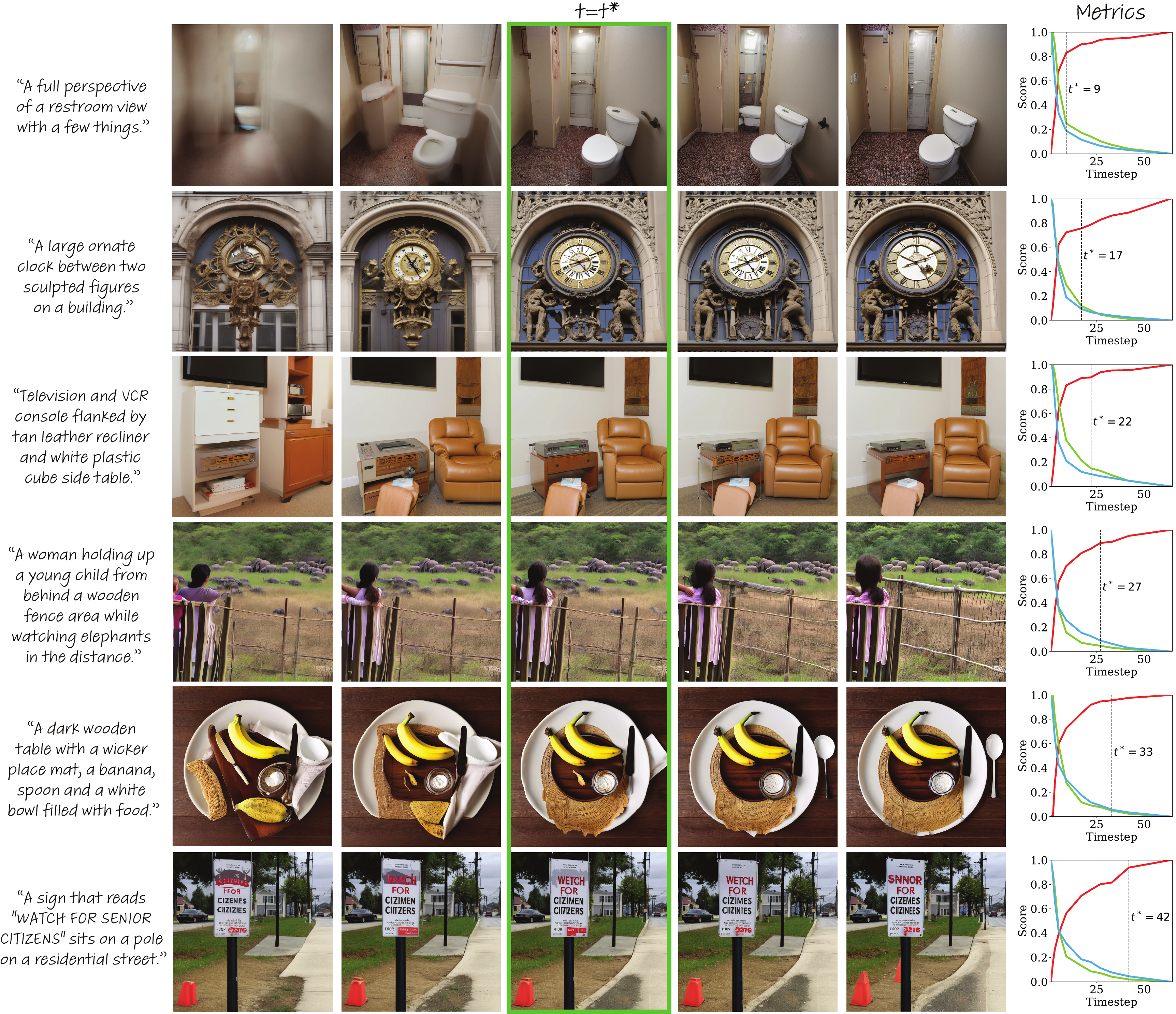}
  \Caption{
      Additional qualitative results.
  }{
    We include qualitative results of a pre-trained diffusion model with images generated at increasing denoising timesteps.
    The center column represents optimal timestep, $\step^*$, predicted with our technique, and with each row having increased $\step^*$.
    The text prompts used to generate the images are in the left column, and the three scores used to compute the optimal timestep are in the right column.
    Green, blue and red curves correspond to metrics scores of L-SNR, D-SIM, and I-CLIP, respectively.    
  }
  \label{fig:supp:addtion}
\end{figure*}

Figure \ref{fig:supp:addtion} provides additional visual comparisons of \methodName-generated images.



\end{document}